\documentclass[journal]{IEEEtran}
\ifCLASSINFOpdf
\else
\fi

\usepackage{times}
\usepackage{epsfig}
\usepackage{graphicx}
\usepackage{amsmath}
\usepackage{dsfont}
\usepackage{amssymb}
\usepackage{url}
\usepackage{amsfonts}
\usepackage{subfigure}
\usepackage{multirow}
\usepackage{makecell}
\usepackage{booktabs}
\usepackage{lineno}

\begin{document}

\title{Beyond Bilinear: Generalized Multimodal Factorized High-order Pooling \\for Visual Question Answering}
\author{Zhou~Yu \IEEEmembership{~Member, IEEE},~
        Jun~Yu \IEEEmembership{~Member, IEEE},~
        Chenchao~Xiang,~
        Jianping~Fan,~
        Dacheng~Tao \IEEEmembership{~Fellow, IEEE}
\thanks{This work was supported in part by National Natural Science Foundation of China under Grant 61702143, Grant 61622205, Grant 61472110 and Grant 61772161, and in part by the Zhejiang Provincial Natural Science Foundation of China under Grant LR15F020002, in part by the Australian Research Council Projects under Grant FL-170100117, Grant LP-150100671 and Grant DP-180103424. (Corresponding author: Jun Yu.)}
\thanks{Z. Yu, J. Yu and C. Xiang are with Key Laboratory of Complex Systems Modeling and Simulation,
School of Computer Science and Technology, Hangzhou Dianzi University, P. R. China (e-mail: yuz@hdu.edu.cn; yujun@hdu.edu.cn; hdu$\_$xcc@163.com).}
\thanks{J. Fan is with Department of Computer Science, University of North Carolina at Charlotte, USA (e-mail: jfan@uncc.edu).}
\thanks{D. Tao is with the UBTECH Sydney Artificial Intelligence Centre and the School of Information Technologies, the Faculty of Engineering and Information Technologies, the University of Sydney, Darlington, NSW 2008, Australia (email: dacheng.tao@sydney.edu.au).}
        }

\markboth{Journal of \LaTeX\ Class Files,~Vol.~14, No.~8, August~2015}%
{Yu \MakeLowercase{\textit{et al.}}: Beyond Bilinear: Generalized Multi-modal Factorized High-order Pooling for Visual Question Answering}
%



\maketitle


\begin{abstract}
Visual question answering (VQA) is challenging because it requires a simultaneous understanding of both visual content of images and textual content of questions. To support the VQA task, we need to find good solutions for the following three issues: 1) fine-grained feature representations for both the image and the question; 2) multi-modal feature fusion that is able to capture the complex interactions between multi-modal features; 3) automatic answer prediction that is able to consider the complex correlations between multiple diverse answers for the same question. For fine-grained image and question representations, a `co-attention' mechanism is developed by using a deep neural network architecture to jointly learn the attentions for both the image and the question, which can allow us to reduce the irrelevant features effectively and obtain more discriminative features for image and question representations. For multi-modal feature fusion, a generalized Multi-modal Factorized High-order pooling approach (MFH) is developed to achieve more effective fusion of multi-modal features by exploiting their correlations sufficiently, which can further result in superior VQA performance as compared with the state-of-the-art approaches. For answer prediction, the KL (Kullback-Leibler) divergence is used as the loss function to achieve precise characterization of the complex correlations between multiple diverse answers with same or similar meaning, which can allow us to achieve faster convergence rate and obtain slightly better accuracy on answer prediction. A deep neural network architecture is designed to integrate all these aforementioned modules into a unified model for achieving superior VQA performance. With an ensemble of our MFH models, we achieve the state-of-the-art performance on the large-scale VQA datasets and win the runner-up in VQA Challenge 2017.
\end{abstract}

\begin{IEEEkeywords}
Visual question answering (VQA), multi-modal feature fusion, co-attention learning, deep learning.
\end{IEEEkeywords}

\section{Introduction}
\IEEEPARstart{T}hanks to recent advances of deep neural networks (DNN) in computer vision and natural language processing, computers are expected to be able to automatically understand the semantics of images and natural languages in the near future. Such advances also continue to redefine and drive research in image-text retrieval \cite{wu2014sparse,yu2014discriminative,shen2017multilabel}, image captioning \cite{donahue2015long,xu2015show}, and visual question answering \cite{antol2015vqa,malinowski2014multi}.

Compared with image-text retrieval and image captioning which just require the underlying algorithms to {search} or {generate} a free-form text description for a given image, \emph{visual question answering} (VQA) is a more challenging task that requires fine-grained understanding of the semantics of both the images and the questions as well as supports complex reasoning to predict the best-matching answer accurately. In some aspects, the VQA task can be treated as a generalization of image captioning and image-text retrieval. Thus building effective VQA algorithms which can achieve performance that is close to that of human beings, is an important step towards the general artificial intelligence.

\begin{figure}
\begin{center}
\includegraphics[width=0.45\textwidth]{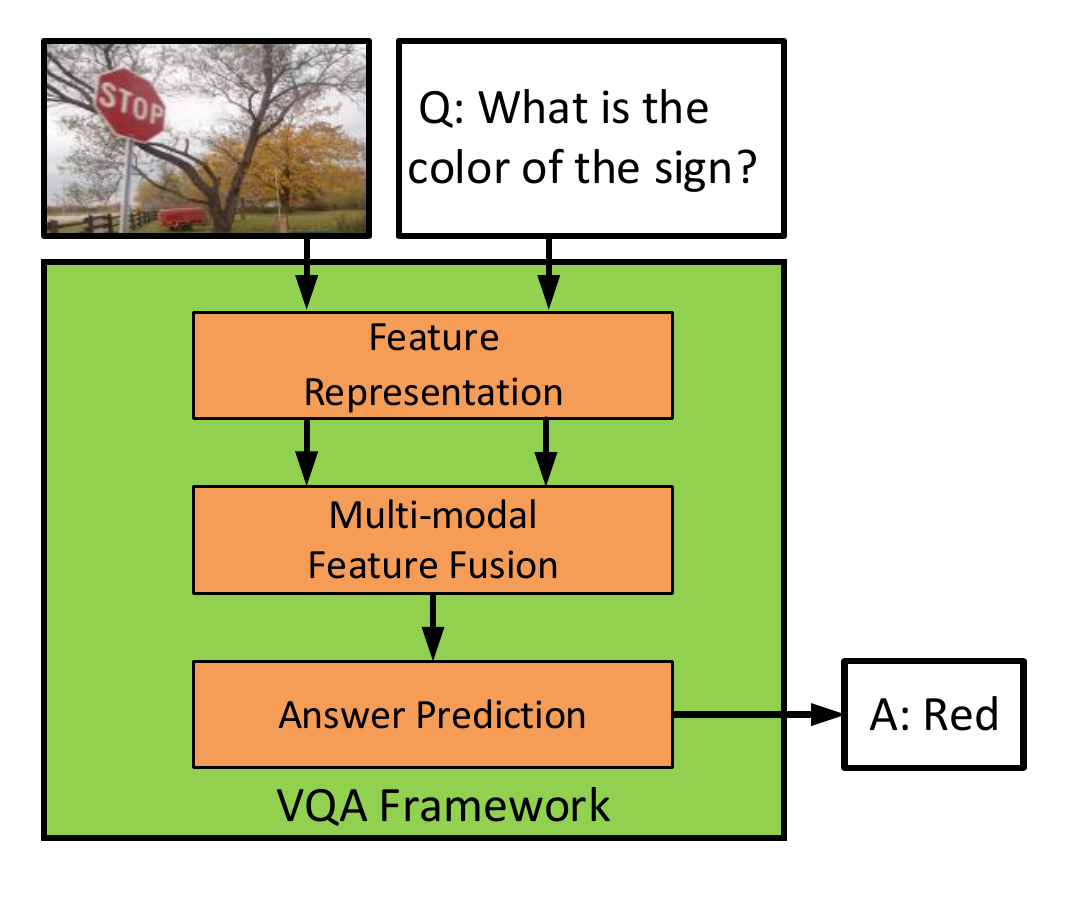}
\caption{The general framework for the VQA task. Given an arbitrary image and an open-vocabulary question (Q) as the inputs, the vqa model outputs the answer (A) in natural language. }
\label{fig:vqa_example}
\end{center}
\vspace{-10pt}
\end{figure}

To support the VQA task, we need to address the following three issues effectively (see the example in Fig. \ref{fig:vqa_example}): (1) extracting discriminative features for image and question representations; (2) combining the visual features from the image and the textual features from the question to generate the fused image-question features; (3) using the fused image-question features to learn a multi-class classifier for predicting the best-matching answer correctly. Deep neural networks (DNNs) are very effective and flexible, most of the existing VQA approaches tackle these three issues in one single DNNs model and train the model in an end-to-end fashion through back-propagation.

For feature-based image representation, directly using the global features extracted from the whole image may introduce noisy information (i.e., irrelevant features) that are irrelevant to the given question, e.g., the given question may strongly relate to only a small part of the image (i.e., image attention region) rather than the whole image. Therefore, it is intuitive to introduce \emph{visual attention} mechanism \cite{xu2015show} into the VQA task to adaptively learn the most relevant image regions for a given question. Modeling visual attention may significantly improve performance \cite{fukui2016multimodal}. On the other hand, the questions interpreted in natural languages may also contain colloquialisms that can be treated as noise, thus it is very important to model the question attention simultaneously. Unfortunately, most existing approaches only model the image attention without considering the question attention. Motivated by these observations, we design a deep network architecture for the VQA task by using a \emph{co-attention learning} module to jointly learn the attentions for both the image and the question, which may allow us to extract more discriminative features for image and question representations.

For multi-modal feature fusion, most existing approaches simply use linear models (e.g., concatenation or element-wise addition) to integrate the visual feature from the image with the textual feature from the question even their distributions may vary dramatically \cite{zhou2015simple,lu2016hierarchical}. Such linear models may not be able to generate expressive image-question features that are able to fully capture the complex correlations between multi-modal features. In contrast to linear pooling, bilinear pooling \cite{tenenbaum1997separating} has recently been used to integrate different CNN features for fine-grained image recognition \cite{lin2015bilinear}. Unfortunately, such bilinear pooling approach may output high-dimensional features for image-question representation and the underlying deep networks for feature extraction may contain huge number of model parameters, which may seriously limit its applicability for VQA. To tackle these problems effectively, Multi-modal Compact Bilinear (MCB) pooling \cite{fukui2016multimodal} and Multi-modal Low-rank Bilinear (MLB) pooling \cite{kim2016hadamard} have been developed to reduce the computational complexity of the original bilinear pooling model and make it practicable for VQA. However, MCB needs very high-dimensional feature to guarantee good performance and MLB needs a great many training iterations to converge to a satisfactory solution. To tackle these problems, we propose a Multi-modal Factorized Bilinear pooling approach (MFB) which enjoys the dual benefits of compact output features of MLB and robust expressive capacity of MCB. Moreover, we extend the bilinear MFB model to a generalized high-order setting and proposed a Multi-modal Factorized High-order pooling (MFH) method to achieve more effective fusion of multi-modal features by exploiting their complex correlations sufficiently. By introducing more complex high-order interactions between multi-modal features, our MFH method can achieve more discriminative image-question representation and further result in significant improvement on the VQA performance.

For answer prediction, some datasets like VQA \cite{antol2015vqa} provide multiple answers for each image-question pair and such diverse answers are typically annotated by different users. As the answers are represented in natural languages, for a given question, different users may provide diverse answers or expressions which have same or similar meaning, thus such diverse answers may have strong correlations and they are not independent at all. For example, both \emph{a little dog} and \emph{a puppy} could be the correct answers for the same question. Motivated by these observations, it is important to design an appropriate mechanism to model the complex correlations between multiple diverse answers for the same question. In MCB \cite{fukui2016multimodal}, an \emph{answer sampling} strategy was proposed to randomly pick an answer from a set of candidates during the training course. In this way, the complex correlations between multiple diverse answers could be eventually learned by the model with sufficient training iterations. In this paper, we formulate the problem of answer prediction as a \emph{label distribution learning} problem. The answers for an image-question pair in the training dataset are converted to a probability distribution over all possible answers. We use the Kullback-Leibler divergence (KLD) as the loss function to achieve more accurate characterization of the consistency between the probability distribution of the predicted answers and the probability distribution of the ground truth answers given by the annotators. Compared with the answer sampling method in MCB \cite{fukui2016multimodal}, using the KLD loss can achieve faster convergence rate and obtain slightly better accuracy on answer prediction.

In summary, we have made the following contributions in this study:
\begin{itemize}
\item A \emph{co-attention} learning architecture is designed to jointly learn the attentions for both the image and the question, which can allow us to reduce the irrelevant features (i.e., noisy information) effectively and obtain more discriminative features for image and question representations.
\item A Multi-modal Factorized Bilinear Pooling (MFB) approach is developed to achieve more effective fusion of the visual features from the image and the textual features from the question. By supporting more effective exploitation of the complex correlations between multi-modal features, our MFB approach can significantly outperform the existing bilinear pooling approaches.
\item A generalized Multi-modal Factorized High-order pooling (MFH) approach is developed by cascading multiple MFB blocks. Compared with MFB, MFH captures more complex correlations of multi-modal feature to achieve more discriminative image-question representation and further result in significant improvement on the VQA performance.
\item The KL divergence (KLD) is used as the loss function to achieve more accurate characterization of the consistency between the predicted answers and the annotated answers, which can allow us to achieve faster convergence rate and obtain slightly better accuracy on answer prediction.
\item Extensive experiments over multiple VQA datasets are conducted to explain the reason why our approaches are effective. Our experimental results demonstrate that: (a) our proposed approaches can achieve the state-of-the-art performance on the real-world VQA datasets; and (b) the normalization techniques are extremely important in bilinear pooling models.
\end{itemize}

The rest of the paper is organized as follows: In section \ref{sec:related_work}, we review the related work of VQA approaches, especially the ones introducing the bilinear pooling. In section \ref{sec:mfb}, we revisit the bilinear model and its factorized extension. Then, we propose the bilinear MFB model and reveal the fact that MFB is a generalization form of MLB. Based on MFB, we further propose its generalized high-order extension MFH. In section \ref{sec:nn_arch}, we propose the co-attention learning network architecture for VQA based on MFB or MFH. In section \ref{sec:answer_corr}, we analyze the importance of modeling answer correlation in VQA and propose a solution with KLD loss. In section \ref{sec:experiments}, we introduce our extensive experimental results for algorithm evaluation and multiple real-word VQA datasets are used to evaluate our proposed approaches. Finally, we conclude this paper in section \ref{sec:conclusion}.

\section{Related Work}\label{sec:related_work}
In this section, we briefly review the most relevant research on VQA, especially those studies that use multi-modal bilinear models.

\subsection{Visual Question Answering (VQA)}
Malinowski \emph{et al.} made an early attempt at solving the VQA task \cite{malinowski2014multi}. Since then, solving the VQA task has received increasing attention from the communities of computer vision and natural language processing. Most existing VQA approaches can be classified into the following three categories: (a) the coarse joint-embedding models \cite{zhou2015simple,antol2015vqa,kim2016multimodal}; (b) the fine-grained joint-embedding models with attention \cite{andreas2016learning,lu2016hierarchical,fukui2016multimodal,ilievski2016focused,nam2016dual}; (c) the external knowledge based models \cite{wang2015explicit,wu2016ask}.

The coarse joint-embedding models are the most straightforward solution for VQA. Image and question are first represented as global features and then integrated to predict the answer. Zhou \emph{et al.} proposed a baseline approach for the VQA task by using the concatenation of the image CNN features and the question BoW (bag-of-words) features, and a linear classifier is learned to predict the answer \cite{zhou2015simple}. Wang \emph{et al.} perform a detailed analysis on the modeling of questions using CNN to obtain better question representations for VQA \cite{wang2017learning}. Some approaches introduce more complex deep models, e.g., LSTM networks \cite{antol2015vqa} or residual networks \cite{kim2016multimodal}, to tackle the VQA task in an end-to-end fashion.

One limitation of joint-embedding models is that their global features may contain noisy information (i.e., irrelevant features), and such noisy global features may not be able to answer the fine-grained problems correctly (e.g., ``what color are the cat's eyes?'') . Therefore, recent VQA approaches introduce the \emph{visual attention} mechanism \cite{xu2015show} into the VQA task by adaptively learning the local fine-grained image features for a given question. Chen \emph{et al.} proposed a question-guided attention map that projects the question embeddings to the visual space and formulates a configurable convolutional kernel to search the image attention region \cite{chen2015abc}. Yang \emph{et al.} proposed a stacked attention network to learn the attention iteratively \cite{yang2016stacked}. Some approaches introduce off-the-shelf object detectors \cite{ilievski2016focused} or object proposals \cite{shih2016look} as the candidates of the attention regions and then use the question to identify the relevant ones. Fukui \emph{et al.} proposed multi-modal compact bilinear pooling to integrate the visual features from the image spatial grids with the textual features from the questions to predict the attention \cite{fukui2016multimodal}. As the VQA task need to fully understand the semantic of the question in natural language, it is necessary to learn the \emph{textual attention} for question simultaneously. Inspired by the works from the NLP community \cite{vaswani2017attention,lin2017structured}, some approaches perform attention learning on both the images and the questions.
Lu \emph{et al.} proposed a co-attention learning framework to alternately learn the image attention and the question attention \cite{lu2016hierarchical}. Nam \emph{et al.} proposed a multi-stage co-attention learning model to refine the attentions based on memory of previous attentions \cite{nam2016dual}.

Despite the joint embedding models can deliver impressive VQA performance, they are not good enough for answering the questions that require complex reasoning or knowledge of common senses. Therefore, introducing external knowledge is beneficial for VQA. However, existing approaches have either only been applied to specific datasets \cite{wang2015explicit}, or have been ineffective on benchmark datasets \cite{wu2016ask}. Thus they still have rooms for further exploration and development.

\subsection{Multi-modal Bilinear Models for VQA}
Multi-modal feature fusion plays a critical and fundamental role in VQA. After the image and the question representations are obtained, concatenation or element-wise summations are most frequently used for multi-modal feature fusion. Since the distributions of two feature sets in different modalities (i.e.,the visual features from images and the textual features from questions) may vary significantly, the representation capacity of the simply-fused features may be insufficient, limiting the final prediction performance.

Fukui \emph{et al.} first introduced the bilinear model to solve the problem of multi-modal feature fusion in VQA \cite{fukui2016multimodal}. In contrast to the aforementioned approaches, they proposed the Multi-modal Compact Bilinear pooling (MCB), which uses the outer product of two feature vectors in different modalities to produce a very high-dimensional feature for quadratic expansion \cite{fukui2016multimodal}. To reduce the computational cost, they used a sampling-based approximation approach that exploits the property that the projection of two vectors can be represented as their convolution. The MCB model outperformed the simple fusion approaches and demonstrated superior performance on the VQA dataset \cite{antol2015vqa}. Nevertheless, MCB usually needs high-dimensional features (e.g., 16,000-D) to guarantee robust performance, which may seriously limit its applicability for VQA due to limitations in GPU memory.

To overcome this problem, Kim \emph{et al.} proposed the Multi-modal Low-rank Bilinear Pooling (MLB) approach based on the Hadamard product of two feature vectors (i.e., the image feature $x\in\mathbb{R}^m$ and the question feature $y\in\mathbb{R}^n$) in the common space with two low-rank projection matrices \cite{kim2016hadamard}:
\begin{equation}\label{eq:mlb}
z = \mathrm{MLB}(x,y) = (U^Tx)\circ (V^Ty)
\end{equation}
where $U\in\mathbb{R}^{m\times o}$ and $V\in\mathbb{R}^{n\times o}$ are the projection matrices, $o$ is the dimensionality of the output feature, and $\circ$ denotes the Hadamard product or the element-wise multiplication of two vectors. To further increase model capacity, nonlinear activation like $tanh$ is added after $z$. Since the MLB approach can generate feature vectors with low dimensions and deep networks with fewer model parameters, it has achieved very comparable performance to MCB. In MLB \cite{kim2016hadamard}, the experimental results indicated that MLB may lead to a slow convergence rate (the MLB with attention model takes 250k iterations with the batch size 200, which is about 140 epochs, to converge \cite{kim2016hadamard}).

\section{Generalized Multi-modal Factorized High-order Pooling}\label{sec:mfb}
In this section, we first revisit the multi-modal bilinear models and then introduce the Multi-modal Factorized Bilinear pooling (MFB) model. We give detailed explanation on the implementation of our MFB model and further analyze its relationship with the existing MLB approach \cite{kim2016hadamard}. By treating our MFB model as the basic building block, we extend the idea of bilinear pooling to a generalized high-order pooling and we further propose a Multi-modal High-order pooling (MFH) model by simply cascading multiple MFB blocks to capture more complex high-order interactions between multi-modal features.

\subsection{Multi-modal Factorized Bilinear Pooling}\label{subsec:mfb}
Given two feature vectors in different modalities, e.g., the visual features $x\in\mathbb{R}^{m}$ for an image and the textual features $y\in\mathbb{R}^{n}$ for a question, the simplest multi-modal bilinear model is defined as follows:
\begin{equation}\label{eq:multimodal_bilinear_base}
z_i = x^TW_iy
\end{equation}
where $W_i\in\mathbb{R}^{m\times n}$ is a projection matrix, $z_i\in\mathbb{R}$ is the output of the bilinear model. The bias term is omitted here since it is implicit in $W$. To obtain a $o$-dimensional output $z$, we need to learn $W=[W_i,...,W_o]\in\mathbb{R}^{m \times n \times o}$. Although bilinear pooling can effectively capture the pairwise interactions between the feature dimensions, it also introduces huge number of parameters that may lead to high computational cost and a risk of over-fitting.

Inspired by the matrix factorization tricks for uni-modal data \cite{li2016factorized,rendle2010factorization,tao2016manifold,tao2016person,tao2016ensemble}, the projection matrix $W_i$ in Eq.(\ref{eq:multimodal_bilinear_base}) can be factorized as two low-rank matrices:
\begin{equation}\label{eq:mfb}
\begin{array}{rcl}
z_i &=& x^TU_iV_i^Ty = \sum\limits_{d=1}^kx^Tu_dv_d^Ty\\
&=& \mathds{1}^T(U_i^Tx\circ V_i^Ty)
\end{array}
\end{equation}
where $k$ is the factor or the latent dimensionality of the factorized matrices $U_i=[u_1,...,u_k]\in\mathbb{R}^{m \times k}$ and $V_i=[v_1,...,v_k]\in\mathbb{R}^{n \times k}$, $\circ$ is the Hadamard product or the element-wise multiplication of two feature vectors, $\mathds{1}\in\mathbb{R}^k$ is an all-one vector.

To obtain the output feature $z\in\mathbb{R}^o$ by Eq.(\ref{eq:mfb}), the weights to be learned are two three-order tensors $U=[U_1,...,U_o]\in\mathbb{R}^{m\times k \times o}$ and $V=[V_1,...,V_o]\in\mathbb{R}^{n\times k \times o}$ accordingly. Without loss of generality, we can reformulate $U$ and $V$ as 2-D matrices $\tilde{U}\in\mathbb{R}^{m \times ko}$ and $\tilde{V}\in\mathbb{R}^{n \times ko}$ respectively with simple reshape operations. Accordingly, Eq.(\ref{eq:mfb}) is rewritten as follows:
\begin{equation}\label{eq:mfb_matrix}
z = \mathrm{SumPool}(\tilde{U}^Tx\circ \tilde{V}^Ty, k)
\end{equation}
where the function $\mathrm{SumPool}(x,k)$ means using a one-dimensional non-overlapped window with the size $k$ to perform sum pooling over $x$. We name this model Multi-modal Factorized Bilinear pooling (MFB).

\begin{figure}
\begin{center}
\subfigure[Multi-modal Factorized Bilinear Pooling] {\includegraphics[width=0.6\linewidth]{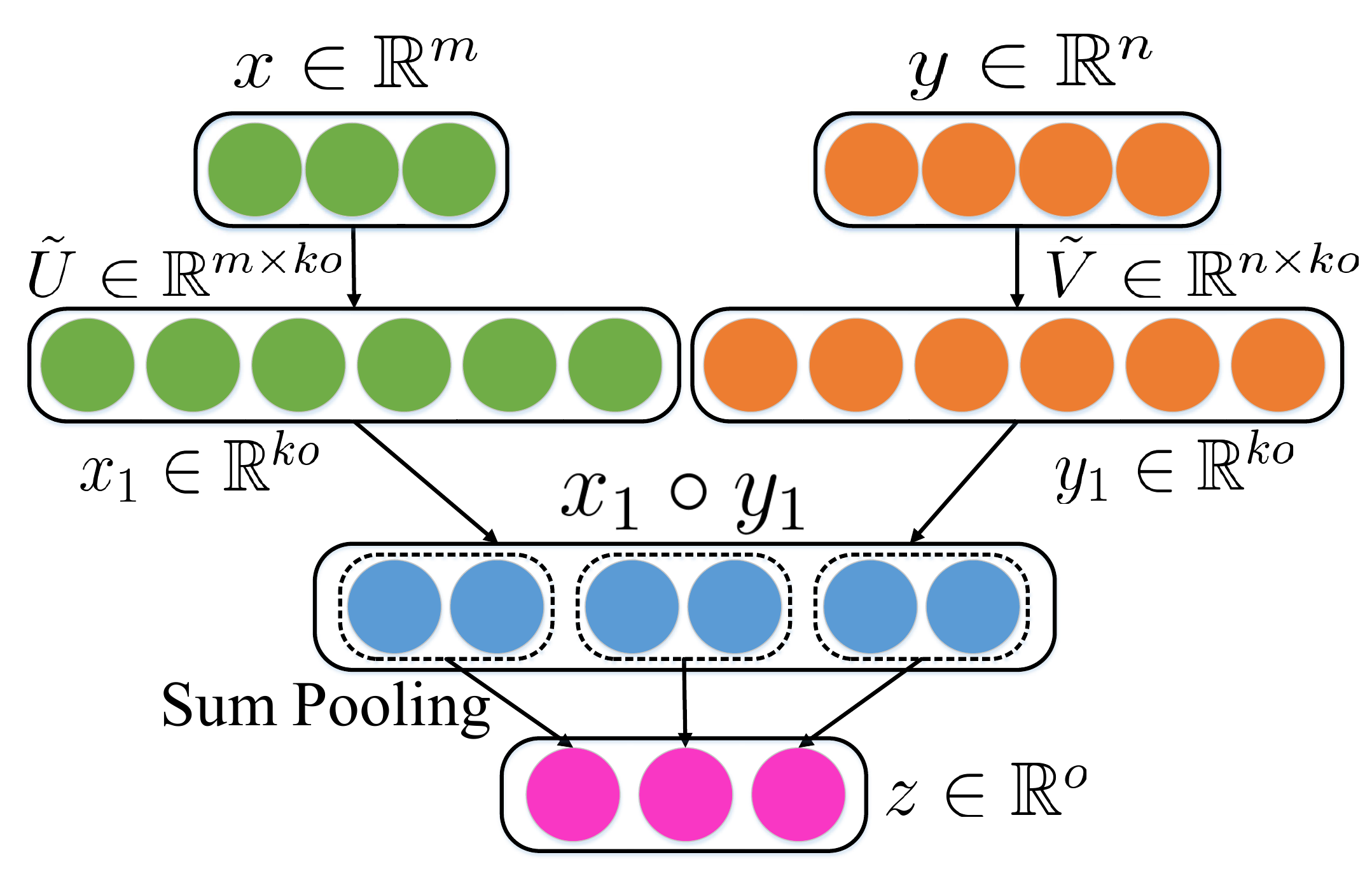}\label{fig:mfb}}
\subfigure[MFB module] {\includegraphics[width=0.35\linewidth]{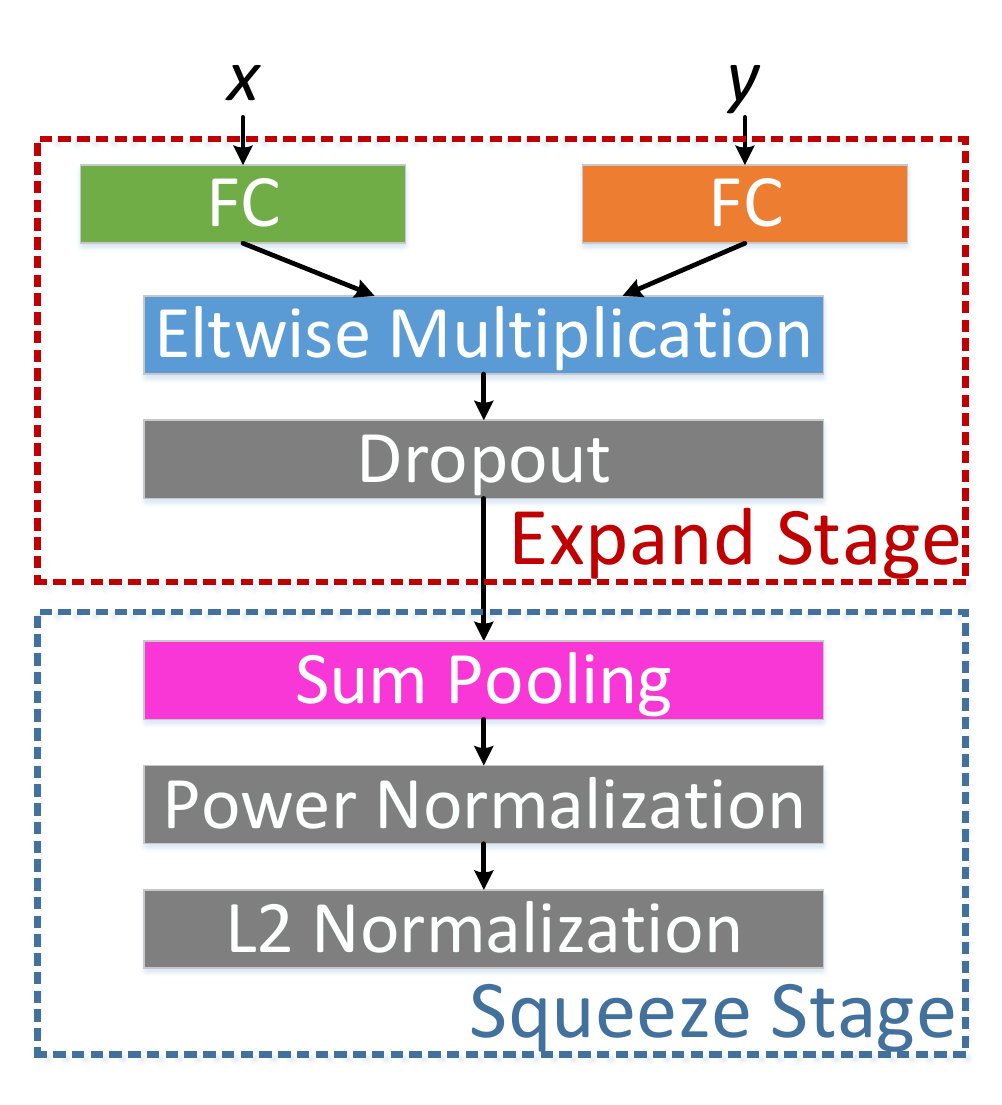}\label{fig:mfb_module}}
\caption{The flowchart of Multi-modal Factorized Bilinear Pooling and complete MFB module.}
\label{fig:mfb_framework}
\end{center}
\vspace{-10pt}
\end{figure}

The detailed procedures of MFB are illustrated in Fig. \ref{fig:mfb}. The approach can be easily implemented by combining some commonly-used layers such as fully-connected, element-wise multiplication and pooling layers. Furthermore, to prevent over-fitting, a dropout layer \cite{srivastava2014dropout,shen2017continuous} is added after the element-wise multiplication layer. Since element-wise multiplication is introduced, the magnitude of the output neurons may vary dramatically, and the model might converge to an unsatisfactory local minimum. Therefore, similar to \cite{fukui2016multimodal}, the power normalization ($z \leftarrow \mathrm{sign}(z)|z|^{0.5}$) and $\ell_2$ normalization ($z \leftarrow z/\|z\|$) layers are appended after MFB output. The flowchart of the entire MFB module is illustrated in Fig. \ref{fig:mfb_module}.

\textbf{Relationship to MLB. }
Eq.(\ref{eq:mfb_matrix}) shows that the MLB in Eq.(\ref{eq:mlb}) is a special case of the proposed MFB with $k=1$, which corresponds to the rank-1 factorization. Figuratively speaking, MFB can be decomposed into two stages (see in Fig. \ref{fig:mfb_module}): first, the features from different modalities are \emph{expanded} to a high-dimensional space and then integrated with element-wise multiplication. After that, sum pooling followed by the normalization layers are performed to \emph{squeeze} the high-dimensional feature into the compact output feature, while MLB directly projects the features to the low-dimensional output space and performs element-wise multiplication. Therefore, with the same dimensionality for the output features, we can conjecture that MLB may suffer from insufficient representation.

\subsection{From Bilinear Pooling to Generalized High-order Pooling}

From the previous work like \cite{fukui2016multimodal,kim2016multimodal}, we have witnessed that the bilinear pooling models have superior representation capacity than the traditional linear pooling models. This inspires us that exploiting the complex interactions among the feature dimensions is beneficial for capturing the common semantics of multi-modal features \cite{shi2017novel,zhao2017beyond}. Therefore, a natural idea is to extend the second-order bilinear pooling to the generalized high-order pooling to further enhance the representation capacity of fused features. In this section, we introduce a generalized Multi-modal Factorized High-order pooling (MFH) model by cascading multiple MFB blocks.

As shown in Fig. \ref{fig:mfb_module}, the MFB module can be separated into the {expand} stage and the {squeeze} stage as follows.
\begin{equation}\label{eq:mfb_exp}
z_{exp} = \mathrm{MFB}_{exp} (x,y) = \mathrm{Dropout}(\tilde{U}^Tx\circ \tilde{V}^Ty)\in\mathbb{R}^{ko}
\end{equation}

\begin{equation}\label{eq:mfb_sqz}
z = \mathrm{MFB}_{sqz} (z_{exp}) = \mathrm{Norm}(\mathrm{SumPool}(z_{exp}))\in\mathbb{R}^{o}
\end{equation}
where $\mathrm{Drop}(\cdot)$, $\mathrm{SumPool}(\cdot)$ and $\mathrm{Norm}(\cdot)$ refer to the dropout, sum pooling and normalization layers respectively. $z_{exp}$ and $z$ are the internal and the output feature of the MFB module respectively.

To make $p$ MFB blocks cascadable, we slightly modify the original MFB$_{exp}$ stage in Eq.(\ref{eq:mfb_exp}) as follows:
\begin{equation}\label{eq:mfh_exp}
\begin{array}{rcl}
z^i_{exp} = \mathrm{MFB}^{i}_{exp}(x,y)= z^{i-1}_{exp} \circ (\mathrm{Dropout}(\tilde{U^i}^Tx\circ \tilde{V^i}^Ty))
\end{array}
\end{equation}
where $i\in\{1,2,...,p\}$ is the index for the MFB blocks. $\tilde{U^i}$, $\tilde{V^i}$ and $z^{i}_{exp}$ are the weight matrices and the internal feature for $i_{th}$ MFB block respectively. $z^{i-1}_{exp}$ are the internal feature of $i-1_{th}$ MFB block and $z_{exp}^{0}\in\mathds{1}^{ko}$ is an all-one vector.

After the internal feature $z^{i}_{exp}$ is obtained for $i$-th MFB block, the output feature $z^{i}$ for $i$-th MFB block can be computed by Eq.(\ref{eq:mfb_sqz}). The final output feature $z$ of the high-order $\mathrm{MFH}^p$ model is obtained by concatenating the output feature of $p$ MFB blocks as follows:
\begin{equation}\label{eq:mfh_p}
z = \mathrm{MFH}^{p} = [z^1,z^2,...,z^p]\in\mathbb{R}^{op}
\end{equation}

\begin{figure}
\begin{center}
\includegraphics[width=0.43\textwidth]{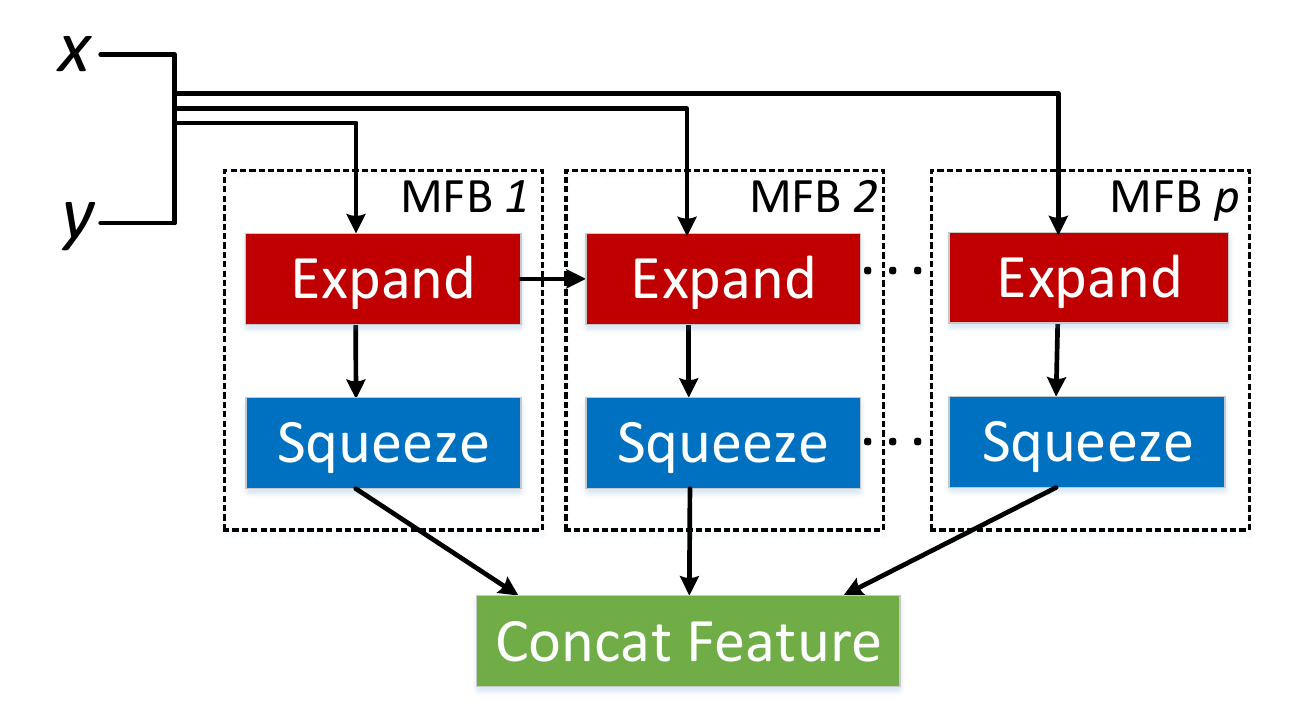}
\caption{Flowchart of the MFH$^{p}$ module based on the cascading of $p$ MFB blocks. Note that MFB is a special case of MFH$^p$ with $p=1$.}
\label{fig:mfhb-flowchart}
\end{center}
\vspace{-10pt}
\end{figure}

The overall flowchart of the MFH approach is illustrated in Fig. \ref{fig:mfhb-flowchart}. With the increase of $p$, the model size and the dimensionality of the output feature for MFH grow linearly. In order to control the model complexity and the training time that we can afford, we use $p<4$ in our experiments. It is worth noting that the propoed MFB model in section \ref{subsec:mfb} is a special case of our MFH$^p$ model with $p=1$.

\section{Network Architectures for VQA}\label{sec:nn_arch}
The goal of the VQA task is to answer a question about an image. The inputs to the model contain an image and a corresponding question about the image. Our model extracts the representations for both the image and the question, integrates multi-modal features by using the MFB or MFH module in Fig. \ref{fig:mfb_module}, treats each individual answer as one class and performs multi-class classification to predict the correct answer. In this section, two network architectures are introduced. The first one is the baseline with one MFB or MFH module, which is used to perform ablation analysis with different hyper-parameters for comparison with other baseline approaches. The second one introduces co-attention learning to achieve more effective characterization of the fine-grained correlations between multi-modal features, which may result in a model with better representation capability.

\subsection{The Baseline Model}
\begin{figure}
\begin{center}
\includegraphics[width=0.49\textwidth]{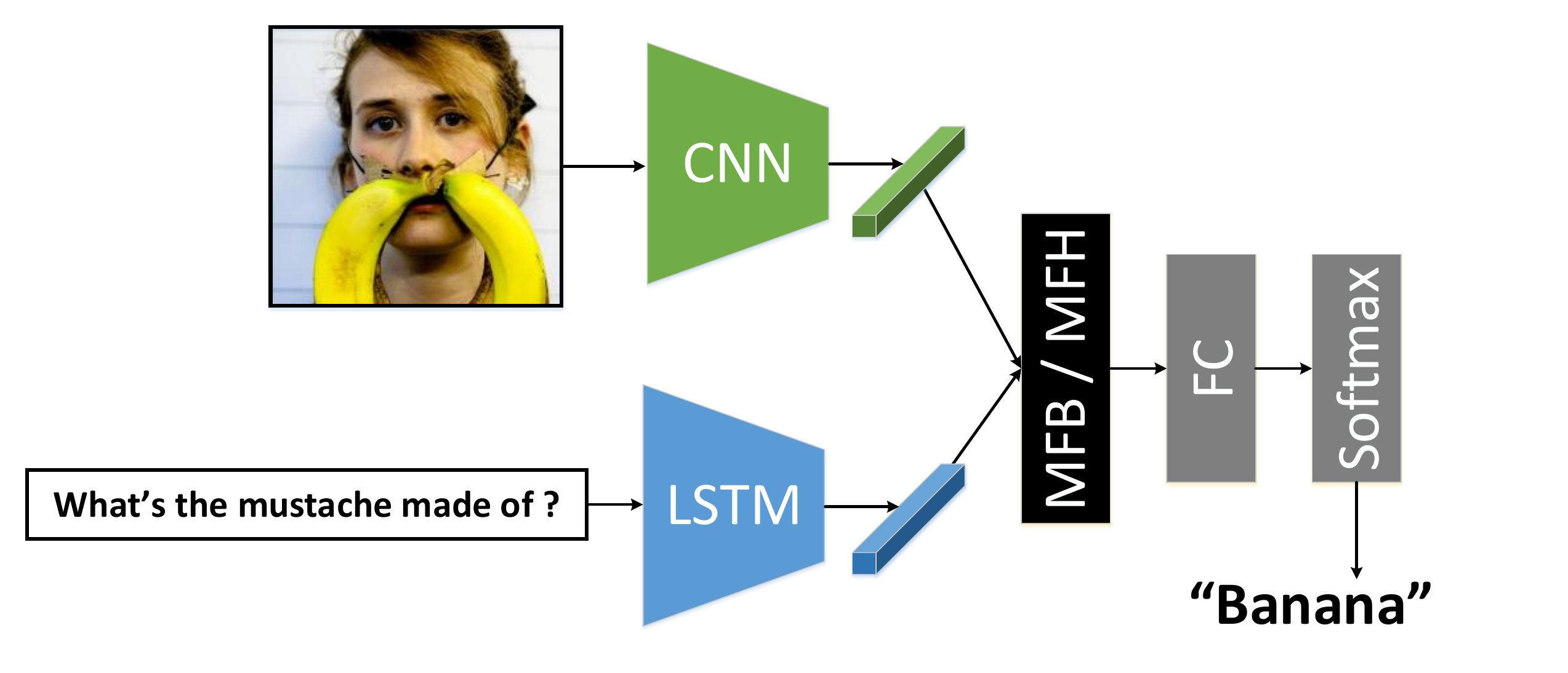}
\caption{The baseline network architecture with MFB or MFH and without the attention mechanism for VQA.}
\label{fig:mfb_baseline}
\end{center}
\end{figure}

Similar to MCB \cite{fukui2016multimodal}, we extract the image features by using 152-layer ResNet model \cite{he2015deep} which is pre-trained on the ImageNet dataset. Images are resized to 448 $\times$ 448, and 2048-D \emph{pool5} features (with $\ell_2$ normalization) are used for image representation. Questions are first tokenized into words, and then further transformed to one-hot feature vectors with max length $T$. Then, the one-hot vectors are passed through an embedding layer and fed into an LSTM networks with 1024 hidden units \cite{hochreiter1997long}. Similar to MCB \cite{fukui2016multimodal}, we extract the output feature of the last word from the LSTM network to form a vector for question representation. For predicting the answers, we simply use the top-$N$ most frequent answers as $N$ classes since they follow the long-tail distribution.

The multi-modal features (that are extracted from the image and the question) are fed to the MFB or MFH module to generate the fused image-question feature $z$. Finally, $z$ is fed to an $N$-way classifier to predict the best-matching answer. Therefore, all the weights except the ones for the ResNet (due to the limitation of GPU memory) are optimized jointly in an end-to-end manner. The whole network architecture is illustrated in Fig. \ref{fig:mfb_baseline}.

\subsection{The Co-Attention Model}
\begin{figure*}
\begin{center}
\includegraphics[width=1\textwidth]{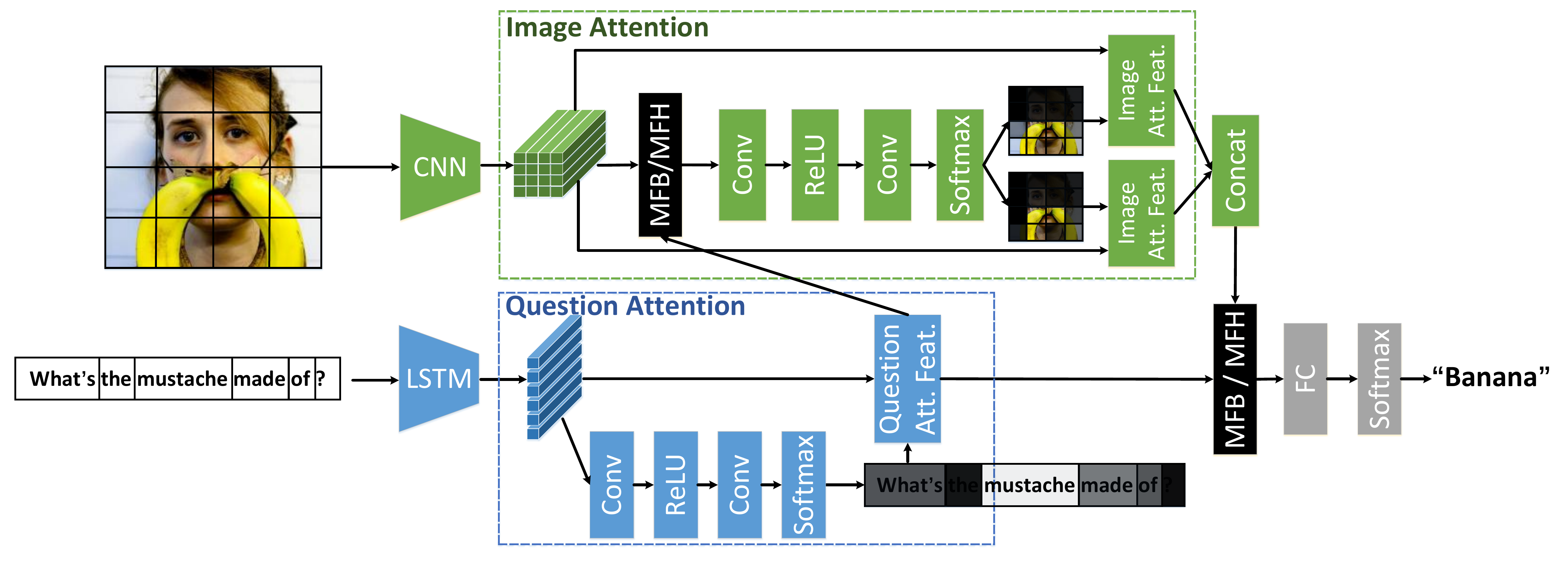}
\caption{The co-attention network architecture with MFB or MFH for VQA. Different from the network of MFB baseline, the images and questions are firstly represented as the fine-grained features respectively. Then, \emph{Question Attention} and \emph{Image Attention} modules are jointly modeled in the framework to provide more accurate answer predictions. For both the image and question attention modules, multiple attention maps (see the example in the image attention module) can be adapted to further improve the representation capacity of the fine-grained features.}
\label{fig:mfb_coatt}
\end{center}
\end{figure*}
For a given image, different questions could result into an entire different set of answers. Therefore, an \emph{image attention} model, which can predict the relevance between each spatial grid of the image with the question, is beneficial for predicting the best-matching answer accurately. From the results reported in MCB \cite{fukui2016multimodal}, one can see that incorporating such image attention mechanism allows the model to effectively learn which image region is important for the question, clearly contributing to better performance than the models without using attention. However, their attention model only focuses on learning the image attention while completely ignoring the question attention. Since the questions are interpreted in natural languages, the contribution of each word is definitely different. Therefore, we develop a co-attention learning approach named MFB+CoAtt or MFH+CoAtt (see Fig. \ref{fig:mfb_coatt}) to jointly learn the attentions for both the question and the image.

Specifically, 14$\times$14 (196) spatial grids of the image (\emph{res5c} feature maps in ResNet) are used to represent the input image and $T$ output features from the LSTM networks are used to represent each word in the input question. After that, the $T$ question features are fed into a \emph{question attention} module and output an attentive question representation. This attentive question representation is fed into an \emph{image attention} module (with 196 image features), and MFB or MFH is used to generate a fused image-question representation. Such fused image-question representation is further used to learn a multi-class classifier for answer prediction. In our excrements, we find that using MFH rather than MFB in the image attention module does not improve the prediction accuracy significantly while inducing much higher computational cost. Therefore, in most of our experiments (unless in the final model ensemble experiment), the MFH module is only used in the feature fusion stage for integrating the attentive features extracted from the image and the question.

Both the image attention module and question attention module consist of sequential 1 $\times$ 1 convolutional layers and ReLU layers followed by the softmax normalization layers to predict the attention weight for each input feature. The attentive feature are obtained by the weighted sum of the input features. To further improve the representation capacity of the attentive feature, multiple attention maps are generated to enhance the learned attention map, and these attention maps are concatenated to output the attentive image features.

It is worth noting that the question attention in our network architecture is learned in a \emph{self-attentive} manner by using the question feature itself. This is different from the image attention module which is learned by using both the image features and question features. The reason is that we assume that the question attention (i.e., the key words of the question) can be inferred without seeing the image, as humans do.

\section{Answer Correlation Modeling}\label{sec:answer_corr}
In most existing VQA approaches, the answering stage is formulated as a multi-class classification problem and each answer refers to an individual class. In practice, this assumption may not hold for the VQA task because the answers with the same or similar meaning can be expressed diversely by different annotators. For example, both the answers `\emph{a little dog}' and `\emph{a puppy}' could be correct for a given image-question pair. Therefore, it is crucial to model the answer correlations in the VQA task so that the learned model could be more robust.

In some datasets like VQA \cite{antol2015vqa}, each question is annotated with multiple answers by different people. To exploit the answer correlations, an \emph{answer sampling} strategy was used in MCB \cite{fukui2016multimodal}. Specifically, for each image-question pair in the training set, the multiple answers for each sample are represented as a distribution vector of all the possible answers $y\in\mathbb{R}^N$, where $N$ is the total number of answers for the whole training set. $y_i\in[0,1]$ indicates the occurrence probability of the $i$-th answer with that $\sum_{i}y_i=1$. In each epoch the sample is accessed, a single answer is obtained by sampling from probability distribution $y$ as the label for this sample in this epoch. In this way, the problem become the traditional multi-class classification problem with single label and traditional softmax loss function could be used to train the model. With sufficient number of iterations, the model can learn the answer correlation eventually.

In practice, using answer sampling strategy may introduce uncertainty to the learned model and may take more iterations to converge. To overcome the problem, we transform the single-label multi-class classification problem with sampled answers to the \emph{label distribution learning} (LDL) problem \cite{geng2013facial} with the answer distribution $y$. Accordingly, we use the KL-divergence loss function to penalize the prediction $z\in\mathbb{R}^N$ after the softmax activation of the last fully-connected layer.
\begin{equation}\label{eq:ldl}
\ell(y,z)_{KL} = \sum\limits_{i}y_i\mathrm{log}({\frac{y_i}{z_i}})
\end{equation}
Note that KL-divergence loss contains an additional constant term compared to the multi-label cross-entropy loss. They are equivalent during optimization.

\section{Experiments}\label{sec:experiments}
We have conducted several experiments to evaluate the performance of our MFB models for the VQA task by using the VQA datasets \cite{antol2015vqa,goyal2016making} to verify our approach. We first perform ablation analysis on the MFB and MFH baseline models to verify the superior performance of the proposed approaches over existing state-of-the-art methods such as MCB \cite{fukui2016multimodal} and MLB \cite{kim2016hadamard}. We then provide detailed analysis of the reasons why our models outperform their counterparts. Finally, we choose the optimal hyper-parameters for the MFB or MFH module and train the models with co-attention for fair comparison with the state-of-the-art approaches on the real-world VQA datasets. The corresponding source codes and pre-trained models are released online\footnote{\url{https://github.com/yuzcccc/vqa-mfb}}.

\subsection{Datasets and Evaluation Criteria}
We have evaluated the performances of our proposed approaches over multiple VQA datasets. In addition, we have compared our proposed approaches with the state-of-the-art algorithms.

\subsubsection{VQA-1.0}
The VQA-1.0 dataset \cite{antol2015vqa} consists of approximately 200,000 images from the MS-COCO dataset \cite{lin2014microsoft}, with 3 questions per image and 10 answers per question. The data set is split into three: \emph{train} (80k images and 240k question-answer pairs), \emph{val} (40k images and 120k question-answer pairs), and \emph{test} (80k images and 240k question-answer pairs). Additionally, there is a 25$\%$ test subset named \emph{test-dev}. Two tasks are provided to evaluate performance: Open-Ended (OE) and Multiple-Choices (MC). We use the tools provided by Antol \emph{et al.} \cite{antol2015vqa} to evaluate the accuracy on the two tasks. Specifically, the accuracy of a predicted answer $a$ is calculated as follows:
\begin{equation}
\mathrm{Accuracy}(a) = \mathrm{min}\left\{\frac{\mathrm{Count}(a)}{3},1 \right\}
\end{equation}
where $\textrm{Count}(a)$ is the count of the answer $a$ voted by different annotators.

\subsubsection{VQA-2.0}
 The VQA-2.0 dataset \cite{goyal2016making} is the updated version of the VQA dataset. Compared with the VQA dataset, it contains more training samples (440k question-answer pairs for training and 214k pairs for validation), and is more balanced to weaken the potential that an overfitted model may achieve good results. Specifically, for every question there are two images in the dataset that result in two different answers to the question. At this point only the train and validation sets are available. Therefore, we report the results of the Open-Ended task on validation set with the model trained on train set. The evaluation criterion on this dataset is same as the one used in the VQA-1.0 dataset.


\subsection{Experimental Setup}
For the VQA and VQA 2.0 datasets, we use the Adam solver with $\beta_1=0.9$, $\beta_2=0.99$. The base learning rate is set to 0.0007 and decays every 40k iterations using an exponential rate of 0.5 for MFB and 0.25 for MFH. All the models are trained up to 100k iterations. Dropouts are used after each LSTM layer (dropout ratio $p=0.3$) and MFB and MFH modules ($p=0.1$). The number of answers $N=3000$. For all experiments (except for the ones shown in Table \ref{table:sota}, which use the train and val sets together as the training set like the comparative approaches), we train on the {train} set, validate on the {val} set, and report the results on the {test-dev} and test-standard sets\footnote{the submission attempts for the test set are strictly limited. Therefore, we report most of our results on the test-dev set and the best results on the test-standard set}. The batch size is set to 200 for the models without the attention mechanism, and set to 64 for the models with attention (due to GPU memory limitation).


All experiments are implemented with the \emph{Caffe} toolbox \cite{jia2014caffe} and performed on the workstations with NVIDIA TitanX GPUs.

\subsection{Ablation Study on the VQA-1.0 Dataset}
We design the following ablation experiments to verify the efficacy of our MFB and MFH modules, as well as the advantage of the KLD loss in modeling answer correlations.

\begin{figure*}
\begin{center}
\subfigure[Standard] {\includegraphics[width=0.24\linewidth]{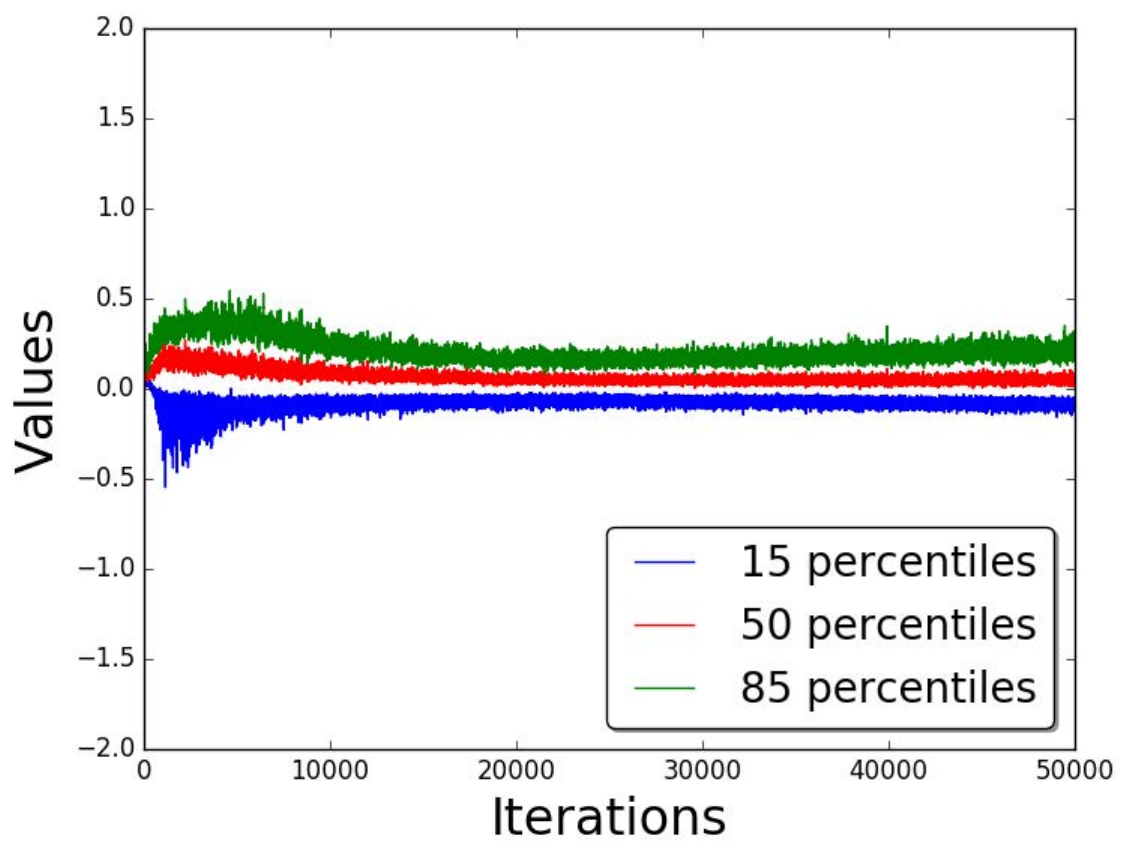}\label{fig:percentile_std}}
\subfigure[w/o power norm.] {\includegraphics[width=0.24\linewidth]{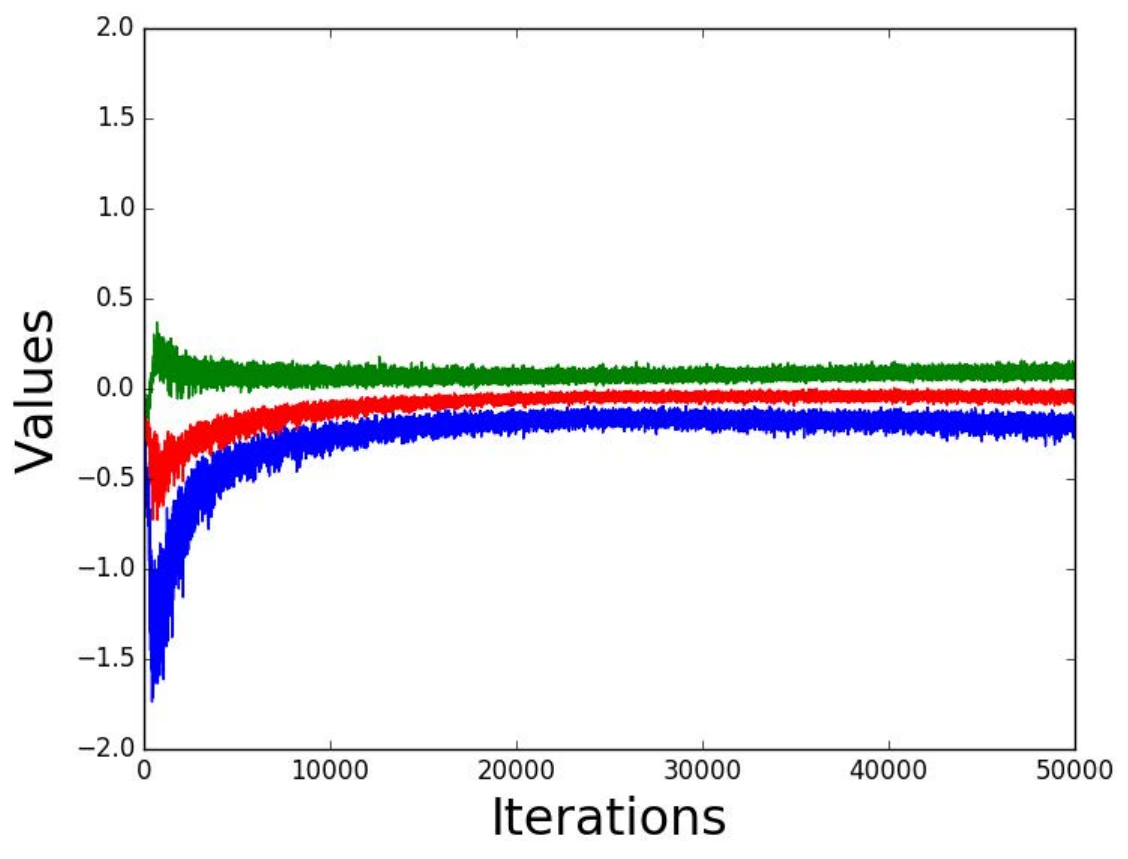}\label{fig:percentile_no_powernorm}}
\subfigure[w/o $\ell_2$ norm.] {\includegraphics[width=0.24\linewidth]{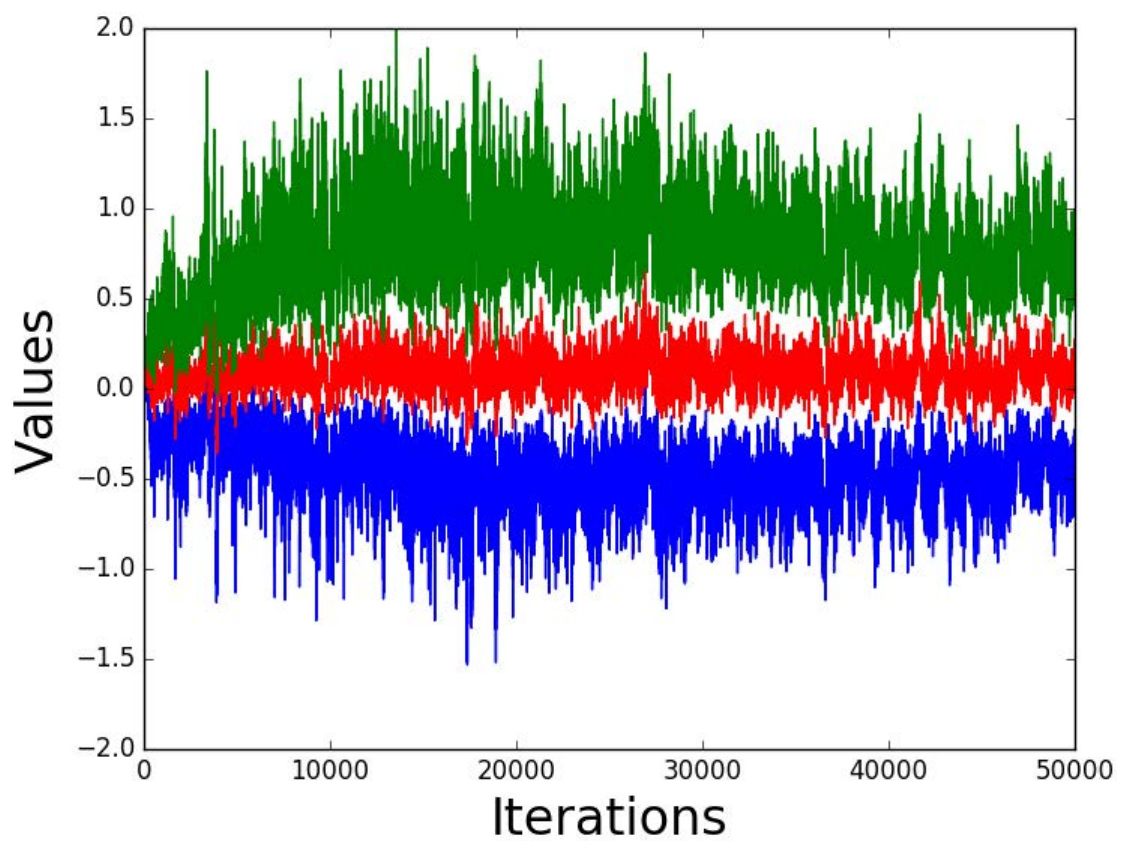}\label{fig:percentile_no_l2norm}}
\subfigure[w/o power and $\ell_2$ norms.] {\includegraphics[width=0.24\linewidth]{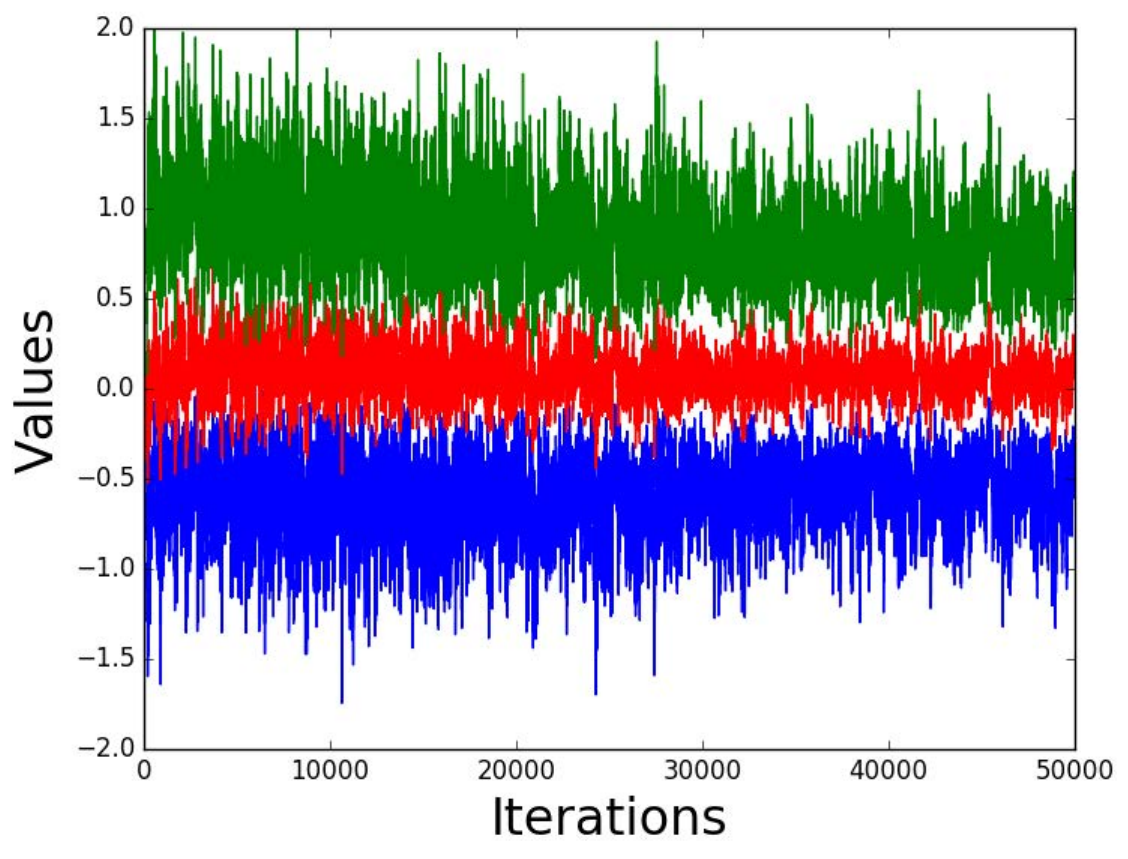}\label{fig:percentile_no_bothnorms}}
\caption{The evolution of the output distribution of one typical neuron with different normalization settings, shown as \{15,50,85\}th percentiles. Both normalization techniques, especially the $\ell_2$ normalization make the neuron values restricted within a narrow range, thus leading to a more stable model. Best viewed in color.}
\label{fig:percentile}
\end{center}
\vspace{-10pt}
\end{figure*}

\subsubsection{Design of the MFB and MFH Module}

In Table \ref{table:base}, we compare the performance of MFB and MFH with other baseline multi-modal fusion models (i.e., feature concatenation, element-wise summation, element-wise product and their variants with one additional fully-connected layer followed by ReLU activation). Besides, the state-of-the-art multi-modal bilinear pooling models, namely MCB \cite{fukui2016multimodal} and MLB \cite{kim2016hadamard} are fairly compared. The models are trained on the train set and evaluated on the test-dev set. For fair comparison, all the compared bilinear pooling approaches use power+$\ell_2$ normalizations. None of these approaches introduce the attention mechanism. We explore different hyper-parameters and normalizations introduced in MFB to explore why MFB outperform the compared bilinear models. Finally, we evaluate MFH$^p$ with different $p$ to explore the effect of high-order feature pooling.

From Table \ref{table:base}, we can see that:

\begin{table}
\centering
\caption{Overall accuracies and model sizes of approaches and on the VQA-1.0 test-dev set of the Open-Ended task. All the compared approaches use the same input features and does not introduce the external datasets or the attention mechanism. The model size includes the parameters for the LSTM networks.
}
\label{table:base}
\begin{tabular}{lccc}
Model & Acc. & Size\\
\hline
Concat  & 57.1 &29M\\
Concat+FC(4096)+ReLU  & 58.4 &45M\\
EltwiseSum  & 56.4 & 23M\\
EltwiseSum+FC(4096)+ReLU  & 58.3 & 37M\\
EltwiseProd  & 57.8 & 23M \\
EltwiseProd+FC(4096)+ReLU  & 58.7 & 37M\\
\hline
MCB \cite{fukui2016multimodal} ($d=16000$) & 59.8 & 63M \\
MLB \cite{kim2016hadamard} ($d=1000$) & 59.7 & 25M \\
\hline
MFB($k=1, o=5000$) & 60.4 & 51M \\
MFB($k=5, o=1000$) & {60.9} & 46M \\
MFB($k=10, o=500$) & 60.5 & 38M \\
\hline
MFB($k=5, o=200$) & 59.8 & 22M \\
MFB($k=5, o=500$) & 60.4 & 28M \\
MFB($k=5, o=2000$) & 60.6 & 62M \\
MFB($k=5, o=4000$) & 60.4 & 107M \\
\hline
MFB($k=5, o=1000$) & - & - \\
~-w/o power norm. & 60.4& - \\
~-w/o $\ell_2$ norm. & 57.7 & -\\
~-w/o power and $\ell_2$ norms. & 57.3 & -\\
\hline
MFH$^2$($k=5,o=1000$)$$ & \textbf{61.6} & 62M\\
MFH$^3$($k=5,o=1000$)$$ & 61.5& 79M
\end{tabular}
\end{table}

First, MFB significantly outperforms all the baseline multi-modal fusion models. MFB is at least 2 points higher than the compared baseline models of similar sizes: the MFB($k$=5,$o$=200) model outperforms the EltwiseProd model by 2.1 points, and the MFB($k$=5,$o$=1000) model outperforms the EltwiseProd+FC+ReLU model by 2.2 points. These results demonstrates the advantage of second-order bilinear pooling models over the first-order pooling models on learning discriminative multi-modal feature representations.

\begin{figure}
\begin{center}
\includegraphics[width=0.45\textwidth]{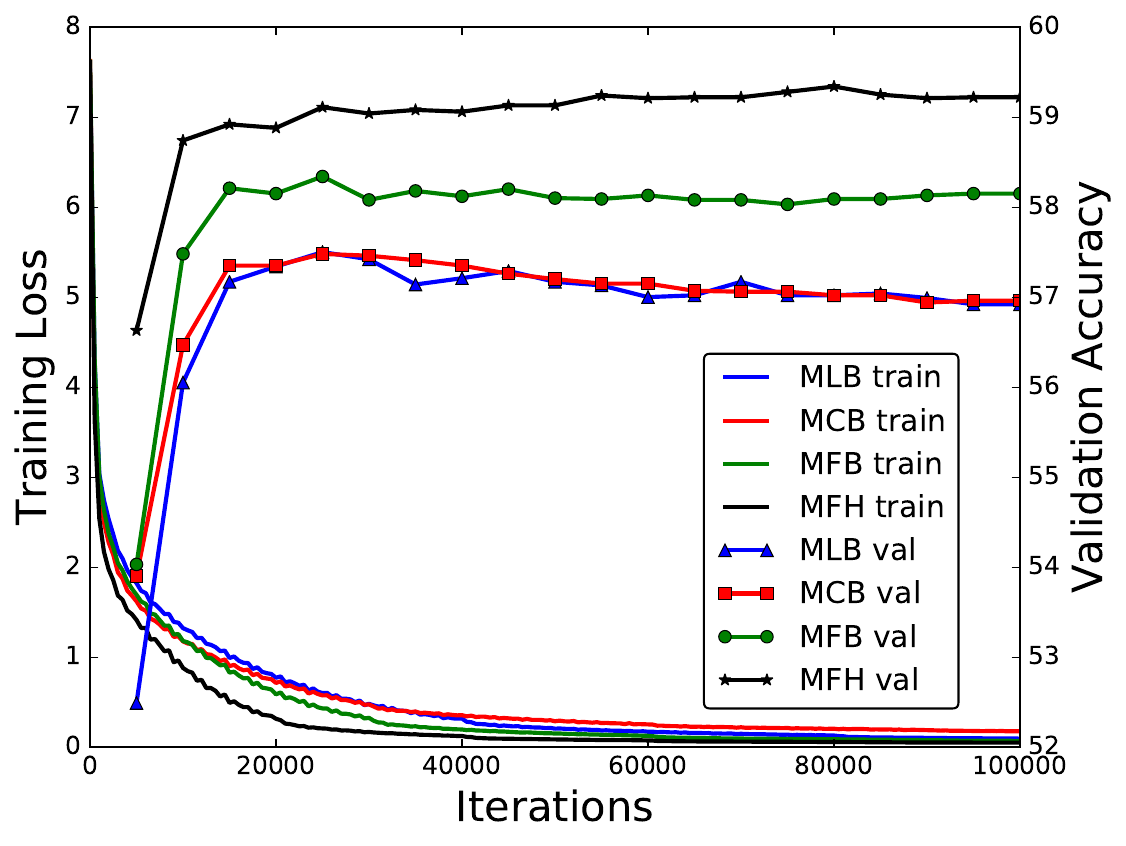}
\caption{The training loss and validation accuracy vs. iterations of MCB, MLB, MFB and MFH$^2$($k=5,o=1000$). KLD loss is used for all the methods. Best viewed in color.}
\label{fig:train_val}
\end{center}
\vspace{-10pt}
\end{figure}

Second, MFB outperforms other multi-modal bilinear pooling approaches. With 5/6 parameters, MFB($k$=5,$o$=1000) achieves an improvement of about 1.0 points compared with MCB and MLB. Moreover, with only 1/3 parameters and 2/3 GPU memory usage, MFB($k$=5,$o$=200) obtains similar results to MCB. These characteristics allows us to train our model on a memory limited GPU with larger batch-size. In Fig. \ref{fig:train_val}, we show the courses of validation, from which it can be seen that
MFB significantly outperforms the two other methods in terms of accuracy on the validation set. Furthermore, it can be seen from the accuracy curve of MCB that its performance gradually falls after 25,000 iterations, indicating that it suffer from overfitting with the high-dimensional output features. In comparison, the performance of MFB is relatively robust.

Third, when $ko$ is fixed to a constant, e.g., 5000, the number of factors $k$ affects the performance. Increasing $k$ from 1 to 5, produces a 0.5 points performance gain. When $k=10$, the performance has approached saturation. This phenomenon can be explained by the fact that a large $k$ corresponds to using a large window to sum pool the features, which can be treated as a compressed representation and may lose some information. When $k$ is fixed, increasing $o$ does not produce any further improvement. This suggests that high-dimensional output features may be easier to overfit. Similar results can be seen in MCB \cite{fukui2016multimodal}. In summary, $k=5$ and $o=1000$ may be a suitable combination for our MFB model on the VQA dataset, so we use these settings in our follow-up experiments.

\begin{figure*}
\begin{center}
\caption{The validation accuracies of MFB and MFB+CoAtt models w.r.t different answer correlation modeling strategies.}
\label{fig:val}
\subfigure[MFB w.r.t. different strategies] {\includegraphics[width=0.4\linewidth]{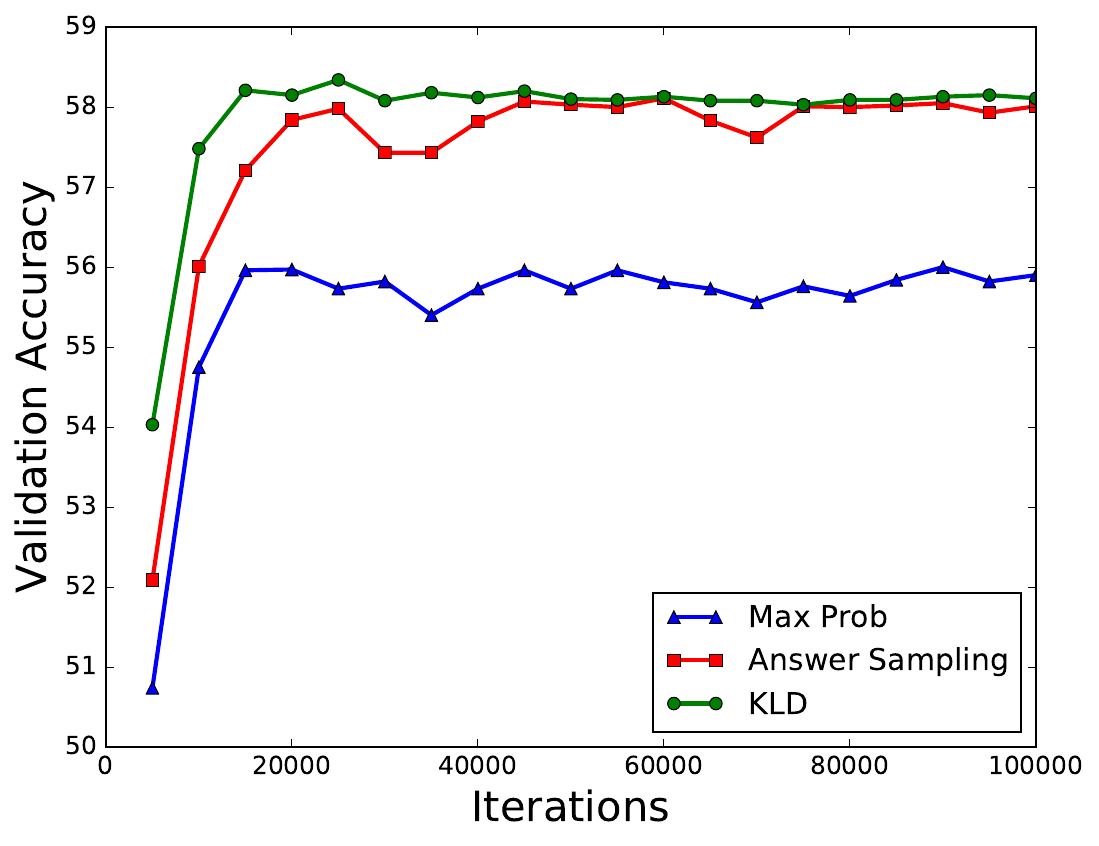}\label{fig:mfb_loss}}
\subfigure[MFB+CoAtt w.r.t. different strategies] {\includegraphics[width=0.4\linewidth]{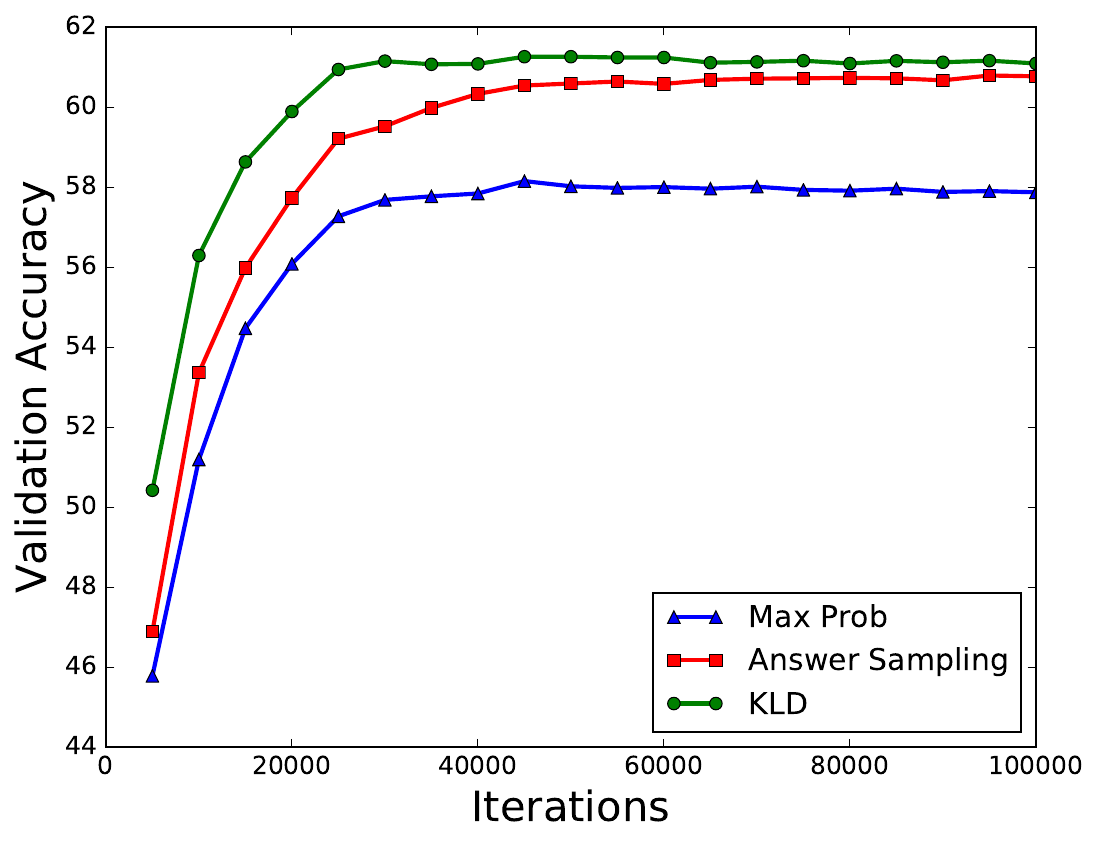}\label{fig:mfb_coatt_loss}}
\end{center}
\vspace{-10pt}
\end{figure*}

Fourth, both the power and $\ell_2$ normalization benefit MFB performance. Power normalization results in an improvement of 0.5 points and $\ell_2$ normalization, perhaps surprisingly, results in an improvement of about 3 points. Results without $\ell_2$ and power normalizations were also reported \cite{antol2015vqa} and are similar to those reported here. To explain why normalization is so important, we randomly choose one typical neuron from the MFB output feature before normalization to illustrate how its distribution evolves over time in Fig. \ref{fig:percentile}. It can be seen that the standard MFB model (with both normalizations) leads to the most stable neuron distribution (i.e. small neuron variance) and without the power normalization, about 10,000 iterations are needed to achieve stabilization. Without the $\ell_2$ normalization, the distribution varies seriously over the entire training course. This observation is consistent with the results shown in Table \ref{table:base}. The effects of power normalization and $\ell_2$ normalization are also observed by \cite{perronnin2010improving}. Furthermore, although MLB does not use any normalization, it introduces the \emph{tanh} activation after the fused feature, which regularizes the distribution of the output feature in some way.

Finally, MFH$^2$ and MFH$^3$ further outperform MFB with an improvement of about 0.7 points on the test-dev set. This observation demonstrates the efficacy of high-order pooling model for VQA. However, the performance of MFH$^3$ is slightly worse than MFH$^2$ even with a more complex model. This may be explained that the representation capacity of MFH is saturated with $p=2$ for the VQA task. Therefore, in our following experiments, $p=2$ is used for MFH and the superscript $p$ is omitted for simplicity.

\subsubsection{Answer Correlation Modeling Strategies}

In Fig. \ref{fig:val}, the validation accuracies of MFB and MFB+CoAtt models w.r.t. different answer sampling strategies are demonstrated respectively. \emph{Max Prob} means using the most frequent answer of the sample as the unique label and formulate the optimization for VQA as the traditional multi-class problem with single label. This strategy refer to the baseline approach that does not consider answer correlation. \emph{Answer Sampling} is the strategy used in MCB \cite{fukui2016multimodal}, which random sample an answer from the candidate answer set at each time. \emph{KLD} is the strategy proposed in section \ref{sec:answer_corr} of this paper.

From the results, we have the following observations. First, modeling answer correlation bring remarkable improvement on the VQA-1.0 dataset. The \emph{Answer Sampling} and \emph{KLD} strategies which model the answer correlation, significantly outperform the \emph{Max Prob} strategy. Second, compared with the \emph{Answer Sampling} strategy, the proposed \emph{KLD} strategy has the merits of faster convergence rate and slightly better accuracy, especially on the complex MFB+CoAtt model.

\subsection{Results on the VQA-1.0 Dataset}\label{sec:res_VQA_1.0}

\begin{table*}
\centering
\caption{Open-Ended (OE) and Multiple-Choice (MC) results on VQA-1.0 dataset compared with the state-of-the-art approaches in terms of accuracy in $\%$. {Att.} indicates whether the approach introduce the attention mechanism explicitly, {W.E.} indicates whether the approach uses external word embedding models. E.D. indicates whether the approach uses external datasets. All the reported results are obtained with \emph{a single model}. For the test-dev set, the best results in each split are bolded. For the test-standard set, the best results overall all the splits are bolded.}
\label{table:sota}
\begin{tabular}{lcccccccccccccccc}
{Model}& {ATT.} & {W.E.} & {E.D.} & \multicolumn{13}{c}{{Accuracy}} \\
\hline
\specialrule{0em}{1pt}{1pt}
&&&& \multicolumn{6}{c}{{Test-dev}} && \multicolumn{6}{c}{{Test-Standard}} \\
\cline{5-10}
\cline{12-17}
\specialrule{0em}{1pt}{1pt}
 &&&&\multicolumn{4}{c}{{OE}} &&{MC} && \multicolumn{4}{c}{{OE}} && {MC}\\
\cline{5-8}
\cline{10-10}
\cline{12-15}
\cline{17-17}
\specialrule{0em}{1pt}{1pt}
 &&&& All & Y/N & Num & Other && All& & All & Y/N & Num & Other && All\\
iBOWIMG \cite{zhou2015simple} &&&& 55.7 & 76.5 & 35.0 & 42.6 && - &&55.9& 78.7& 36.0& 43.4 && 62.0\\
DPPnet \cite{noh2016image} &&&& 57.2 & 80.7 & 37.2 & 41.7 & &- & &57.4& 80.3& 36.9& 42.2 && - \\
VQA team \cite{antol2015vqa} &&&& 57.8 & 80.5 & 36.8 & 43.1 & &62.7& &58.2& 80.6& 36.5& 43.7&& 63.1\\
AYN \cite{malinowski2015ask}    &&&& 58.4 & 78.4 & 36.4 & 46.3 && - && 58.4& 78.2& 36.3& 46.3&& -\\
AMA \cite{wu2016ask}   & &&& 59.2 & 81.0 & 38.4 & 45.2 && - &&59.4& 81.1& 37.1& 45.8&& -\\
MCB \cite{fukui2016multimodal}  & && & 61.1 & 81.7 & 36.9 & 49.0 && - &&61.1& 81.7& 36.9& 49.0&& -\\
MRN \cite{kim2016multimodal}  & && & 61.7 & {82.3} & \textbf{38.9} & 49.3 && - && 61.8& 82.4& 38.2& 49.4&& 66.3 \\
MFB (Ours) &&&& {62.2}& 81.8 & 36.7 & {51.2}&& {67.2} &&-&-&-&-&&-\\
MFH (Ours) &&&& \textbf{62.9} &\textbf{83.1}&36.8&\textbf{51.5}&&\textbf{67.9}&-&-&-&&-\\
\hline
SMem \cite{xu2016ask}  &$\checkmark$& && 58.0 & 80.9 & 37.3 & 43.1 && - &&58.2& 80.9& 37.3& 43.1&&-\\
NMN \cite{andreas2016neural}   & $\checkmark$& && 58.6 & 81.2 & 38.0 & 44.0 && - &&58.7& 81.2& 37.7& 44.0&&-\\
SAN \cite{yang2016stacked}  & $\checkmark$& & & 58.7 & 79.3 & 36.6 & 46.1 && - && 58.9 &-&-&-&&-\\
FDA \cite{ilievski2016focused}   & $\checkmark$&& & 59.2 & 81.1 & 36.2 & 45.8 && -&& 59.5&-&-&-&&-\\
DNMN \cite{andreas2016learning} & $\checkmark$& & & 59.4 & 81.1 & 38.6 & 45.4 && - &&59.4&-&-&-&&-\\
HieCoAtt \cite{lu2016hierarchical}& $\checkmark$&&& 61.8& 79.7 & 38.7 & 51.7 & &65.8 &&62.1&-&-&-&&-\\
RAU \cite{noh2016training} & $\checkmark$&&& 63.3 & 81.9 & 39.0 & 53.0 & &67.7 &&63.2& 81.7& 38.2& 52.8&& 67.3\\
MCB+Att \cite{fukui2016multimodal} & $\checkmark$&&& 64.2& 82.2 & 37.7 & 54.8 && - &&-&-&-&-&&-\\
DAN \cite{nam2016dual} & $\checkmark$&&& 64.3& 83.0 & \textbf{39.1} & 53.9 && 69.1 && 64.2& 82.8& 38.1& 54.0&&69.0\\
MFB+Att (Ours) & $\checkmark$& && 64.6 & 82.5& {38.3}& 55.2&& 69.6 &&-&-&-&-&&-\\
MFB+CoAtt (Ours)  & $\checkmark$& && {65.1}& {83.2} & 38.8& {55.5}&& {70.0} && -&-&-&-&&-\\
MFH+CoAtt (Ours)  & $\checkmark$& && \textbf{65.8}& \textbf{84.1}&{38.1}&\textbf{56.5}& &\textbf{70.6}&&-&-&-&-&&-\\
\hline
MCB+Att+GloVe \cite{fukui2016multimodal} & $\checkmark$& $\checkmark$&& 64.7 & 82.5 & 37.6 & {55.6}&& - &&-&-&-&-&&-\\
MLB+Att+StV \cite{kim2016hadamard} & $\checkmark$& $\checkmark$ && 65.1 & {84.1} & 38.2 & 54.9 &&- && 65.1& 84.0& 37.9& 54.8&& 68.9 \\
MFB+CoAtt+GloVe (Ours) & $\checkmark$& $\checkmark$ && {65.9} & {84.0} & \textbf{39.8} & {56.2}&& {70.6} &&{65.8}& {83.8}& {38.9}& {56.3}&& {70.5}\\
MFH+CoAtt+GloVe (Ours) & $\checkmark$& $\checkmark$ && \textbf{66.8} & \textbf{85.0}& 39.7&\textbf{57.4}&&\textbf{71.4}&&   {66.9}&{85.0}&\textbf{39.5}&{57.4}&&{71.5}\\
\hline
MCB+Att+GloVe+VG \cite{fukui2016multimodal} &$\checkmark$&$\checkmark$&$\checkmark$& 65.4& 82.3& 37.2& 57.4& &-&&-&-&-&-&&-\\
MLB+Att+StV+VG \cite{kim2016hadamard} &$\checkmark$&$\checkmark$&$\checkmark$& 65.8& 83.9& 37.9& 56.8& &-&&-&-&-&-&&-\\
MFB+CoAtt+GloVe+VG (Ours) & $\checkmark$& $\checkmark$ &$\checkmark$& 66.9& 84.1& 39.1& 58.4& &71.3&&66.6&84.2&38.1&57.8&&71.4\\
MFH+CoAtt+GloVe+VG (Ours) & $\checkmark$& $\checkmark$ &$\checkmark$& \textbf{67.7}&\textbf{84.9}&\textbf{40.2}&\textbf{59.2} && \textbf{72.3}&&\textbf{67.5}&\textbf{84.9}&{39.3}&\textbf{58.7}&&\textbf{72.1}\\
\end{tabular}
\end{table*}

\begin{figure*}
\begin{center}
\caption{Typical examples of the learned image and question of the MFB+CoAtt+GloVe model on the VQA-1.0 dataset. The top row shows four examples of four correct predictions while the bottom row shows four incorrect predictions. For each example, the query image, question (Q), answer (A) and prediction (P) are presented from top to bottom; the learned image and question attentions are presented next to them. The brightness of images and darkness of words represent their attention weights.}
\label{fig:mfb_visualization}
\includegraphics[width=1\textwidth]{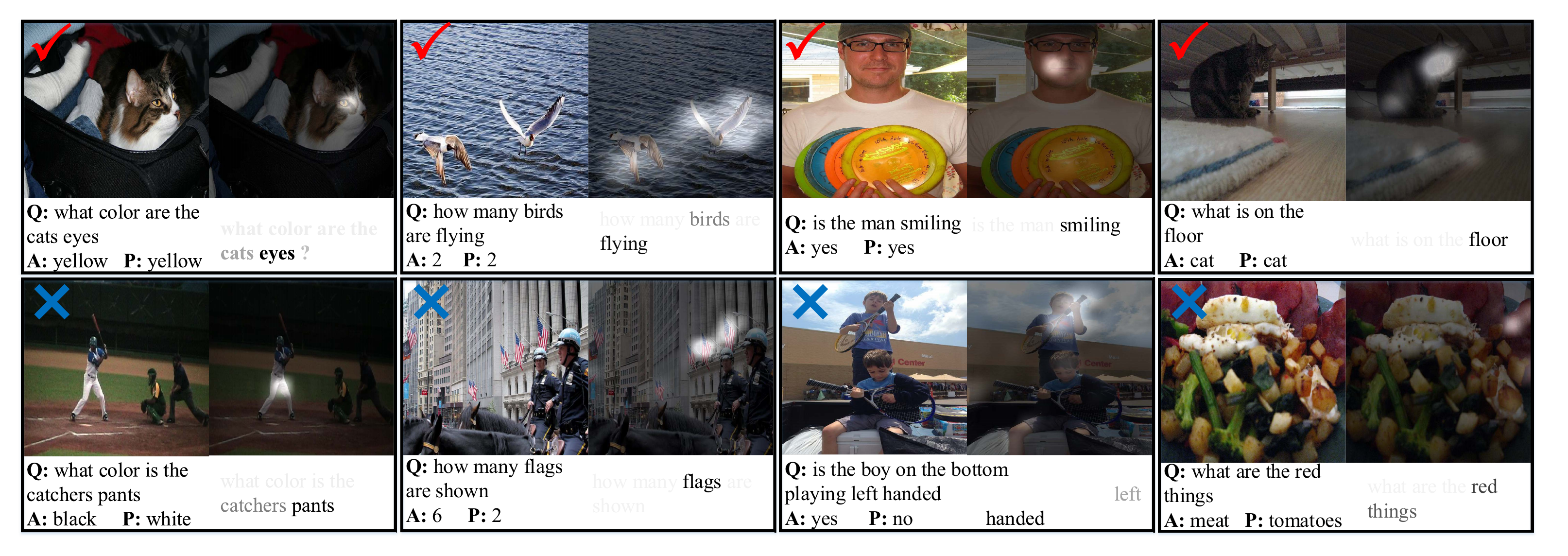}
\end{center}
\end{figure*}

Table \ref{table:sota} compares our approaches with the current state-of-the-art. The table is split into four parts over the rows: the first summarizes the methods without introducing the attention mechanism; the second includes the methods with attention; the third illustrates the results of approaches with external pre-trained word embedding models, e.g., GloVe \cite{pennington2014glove} or Skip-thought Vectors (StV) \cite{kiros2015skip}; and the last includes the models trained with the external large-scale Visual Genome dataset \cite{krishna2016visual} additionally.  To best utilize model capacity, the training data set is augmented so that both the train and val sets are used as the training set. Also, to better understand the question semantics, pre-trained GloVe word vectors are concatenated with the learned word embedding. The MFB model corresponds to the MFB baseline model. The MFB+Att model indicates the model that replaces the MCB with our MFB in the MCB+Att model \cite{fukui2016multimodal}. The MFB+CoAtt model represents the network shown in Fig. \ref{fig:mfb_coatt}. The MFB+CoAtt+GloVe model additionally concatenates the learned word embedding with the pre-trained GloVe vectors. The MFB+CoAtt+GloVe+VG model further introduce the data from the Visual Genome dataset \cite{krishna2016visual} into the training set.

From Table \ref{table:sota}, we have the following observations.

First, the model with MFB outperforms other comparative approaches significantly. The MFB baseline outperforms all other existing approaches without the attention mechanism for both the OE and MC tasks, and even surpasses some approaches with attention. When attention is introduced, MFB+Att consistently outperforms current next-best model MCB+Att, highlighting the efficacy and robustness of the proposed MFB.

Second, the co-attention model further improve the performance over the attention model with only considering the image attention. By additionally introducing the self-attention module for questions, MFB+CoAtt delivers an improvement of 0.5 points on the OE task compared to the MFB+Att model in terms of overall accuracy. Moreover, for each question type (i.e., Y/N, Num or Others), the improvement of MFB+CoAtt over MFB+Att is significant, indicating the effect of the self-attention module in our co-attention learning framework.

Third, by replacing MFB with MFH, the performance of all of our models further enjoy an improvement of about 0.7$\sim$1.1 points steadily. The performance of a single MFH+CoAtt+GloVe model has even surpassed the best published results with an ensemble of 7 MLB or MFB models shown in Table \ref{table:test-std} on the test-standard set.

Finally, with external pre-trained GloVe model and the Visual Genome dataset, the performance of our models are further improved. The MFH+CoAtt+GloVe+VG model significantly outperforms the best reported results with a single model on both the OE and MC task.

In Table \ref{table:test-std}, we compare our model with the state-of-the-art results with model ensemble. Similar with \cite{fukui2016multimodal,kim2016hadamard}, we train 7 individual MFB (or MFH)+CoAtt+GloVe models and average the prediction scores of them. 4 of the 7 models additionally introduce the Visual Genome dataset \cite{krishna2016visual} into the training set. All the reported results are fetched from the leaderboard of the VQA-1.0 dataset\footnote{the \emph{Standard} tab in \url{http://www.visualqa.org/roe.html}}. For fair comparison, only the published results are demonstrated. From the results, the ensemble of MFB models outperforms the next best result by 1.5 points on the OE task and by 2.2 points on the MC task respectively. Furthermore, the result of the ensemble of MFH models obtain a further improvement of 0.8 points and achieve the new state-of-the-art. Finally, compared with the results obtained by human, there is still a lot of room for improvement to approach the human-level.

\begin{table}
\centering
\caption{Comparison with the state-of-the-art results (with model ensemble) on the test-standard set of the VQA-1.0 dataset. Only the published results are demonstrated. The best results are bolded.}
\label{table:test-std}
\begin{tabular}{lccccc}
{Model} & \multicolumn{4}{c}{{OE}} & {MC}\\
\hline
 & All & Y/N & Num & Other & All \\
\cline{2-5}
HieCoAtt \cite{lu2016hierarchical} & 62.1& 80.0& 38.2& 52.0 & 66.1\\
RAU \cite{noh2016training} & 64.1 & 83.3 & 38.0 & 53.4 & 68.1\\
7 MCB models \cite{fukui2016multimodal} & 66.5 & 83.2& 39.5& 58.0& 70.1 \\
7 MLB models \cite{kim2016hadamard} & 66.9 & 84.6& 39.1 &57.8 & 70.3\\
\hline
7 MFB models & {68.4} & {85.6} & {41.0} & {59.8} & {72.5} \\
7 MFH models & \textbf{69.2} & \textbf{86.2} & \textbf{41.8} & \textbf{60.7} & \textbf{73.4}\\
\hline
Human \cite{antol2015vqa} & 83.3& 95.8& 83.4& 72.7& 91.5
\end{tabular}
\end{table}

To demonstrate the effects of co-attention learning, we visualize the learned question and image attentions of some image-question pairs from the {val} set in Fig. \ref{fig:mfb_visualization}. The examples are randomly picked from different question types. It can seen that the learned question and image attentions are usually closely focus on the key words and the most relevant image regions. From the incorrect examples, we can also draw conclusions about the weakness of our approach, which are perhaps common to all VQA approaches: 1) some key words in the question are neglected by the question attention module, which seriously affects the learned image attention and final predictions (e.g., the word \emph{catcher} in the first example and the word \emph{bottom} in the third example); 2) even the intention of the question is well understood, some visual contents are still unrecognized (e.g., the \emph{flags} in the second example) or misclassified (the \emph{meat} in the fourth example), leading to the wrong answer for the counting problem. These observations are useful to guide further improvement for the VQA task in the future.

\subsection{Results on the VQA-2.0 Dataset}
Table \ref{table:vqa2.0} demonstrates our results on the VQA-2.0 dataset ({a.k.a}, VQA challenge 2017). We compare our models with the results of baseline models (including the MCB model which is the champion of VQA Challenge 2016) and the results of the top-ranked teams on the leaderboard. We use the same training strategies aforementioned for this dataset.

From the results, our single MFB and MFH models (with CoAtt+GloVe but without the Visual Genome data argumentation) significantly surpass all the baseline approaches. If we neglect the tiny difference between the results on test-dev and test-standard sets, MFB and MFH is about 2.7 points and 3.5 points higher than the MCB model respectively. Finally, with an ensemble 9 models, we report the accuracy of 68.02$\%$ on the test-dev set and 68.16$\%$ on the test-challenge set respectively \footnote{\url{http://visualqa.org/roe_2017.html}}, which ranks the second place (tied with another team) in VQA Challenge 2017. The details of the 9 models are illustrated in Table \ref{table:mfh_hyperparam}.

In the solution of the champion team, they introduce the region-based visual features extracted from the Faster R-CNN model which is pre-trained on the large-scale Visual Genome dataset \cite{anderson2017up-down}. Using these visual features instead of the convolutional features from the ResNet model brings surprisingly good performance even with a simple VQA model. By using their visual features as the backbone for our models with MFH, we are in the first place on the real-time leaderboard of the VQA-2.0 dataset up to now (15 March, 2018). We report the overall accuracy 70.92$\%$ on the test-standard set of VQA-2.0 with 8 models while they report the accuracy 70.34$\%$ with up to 30 models \cite{teney2017tips}.
\begin{table}
\centering
\caption{The overall accuracies on the test-dev and test-challenge sets of the VQA-2.0 dataset}
\label{table:vqa2.0}
\begin{tabular}{lccccc}

Model & Test-Dev & Test-Challenge \\
\hline
vqateam-Prior & -& 25.98 \\
vqateam-Language &-&	44.34 \\
vqateam-LSTM-CNN &-&	54.08 \\
vqateam-MCB &-&	62.33 \\
Adelaide-ACRV-MSR \cite{teney2017tips} (1st place) &-& \textbf{69.00} \\
DLAIT (2nd place) &-& 68.07 \\
LV$\_$NUS (4th place) &-& {67.62} \\
\hline
1 MFB model & 64.98& - & \\
1 MFH model & 65.80& - & \\
7 MFB models  &67.24& - \\
7 MFH models  &67.96& - \\
9 MFH models (2nd place) &\textbf{68.02}& {68.16}
\end{tabular}
\end{table}

\begin{table}
\centering
\caption{The overall accuracy on the test-dev set of the VQA-2.0 dataset. MFH models with different hyper-parameters are reported. VG indicates whether the training set is augmented with Visual Genome. MFH / MFB(I) means whether the MFH or MFB module is used in the image attention module; $\#$ Q$_{\mathrm{att}}$ and $\#$ I$_{\mathrm{att}}$ indicate the number of glimpses (i.e, attention maps) for the question and image attention modules respectively. }
\label{table:mfh_hyperparam}
\begin{tabular}{ccccc|c}
index &VG &MFH / MFB(I) & {$\#$ Q$_{\mathrm{att}}$} & {$\#$ I$_{\mathrm{att}}$} &  Accuracy($\%$) \\
\hline
1&&MFB&1&2& 65.70\\
2&&MFB&2&2& 65.74\\
3&&MFB&2&3&65.80\\
4&$\checkmark$&MFB&1&2& 65.95\\
5&$\checkmark$&MFB&2&2& 66.12\\
6&$\checkmark$&MFB&2&3 & 66.01\\
7&$\checkmark$&MFH&1&2& 65.93\\
8&$\checkmark$&MFH&2&2& 66.12\\
9&$\checkmark$&MFH&2&3& 66.03\\
\end{tabular}
\end{table}

\section{Conclusions}\label{sec:conclusion}
In this paper, a network architecture with co-attention learning is designed to model both the image attention and the question attention simultaneously, so that we can reduce the irrelevant features effectively and extract more discriminative features for image and question representations. A Multi-modal Factorized Bilinear pooling (MFB) approach is developed to achieve more effective fusion of the visual features from the images and the textual features from the questions, and a generalized high-order model called MFH is developed to capture more complex interactions between multi-modal features. Compared with the existing bilinear pooling methods, our proposed MFB and MFH approaches can achieve significant improvement on the VQA performance because they can achieve more effective exploitation of the complex correlations between multi-modal features. By using the KL divergence as the loss function, our proposed answer prediction approach can achieve faster convergence rate and obtain better performance as compared with the state-of-the-art strategies. Our experimental results have demonstrated that our approaches have achieved the state-of-the-art or comparable performance on two large-scale real-world VQA datasets.

\ifCLASSOPTIONcaptionsoff
  \newpage
\fi



\bibliographystyle{IEEEtran}
\bibliography{TNNLS-2017-P-8216}

\begin{IEEEbiography}[{\includegraphics[width=1in,height=1.25in,clip,keepaspectratio]{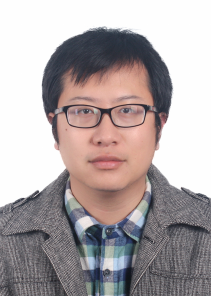}}]{Zhou Yu}
received the B.Eng. and Ph.D. degrees from Zhejiang University, Zhejiang, China, in 2010 and 2015, respectively
He is currently a Lecturer with the School of Computer Science and Technology, Hangzhou Dianzi University, his research interests includes multimodal data analysis, computer vision, machine learning and deep learning.
\end{IEEEbiography}
\vspace{-10pt}
\begin{IEEEbiography}[{\includegraphics[width=1.2in,height=1.25in,clip,keepaspectratio]{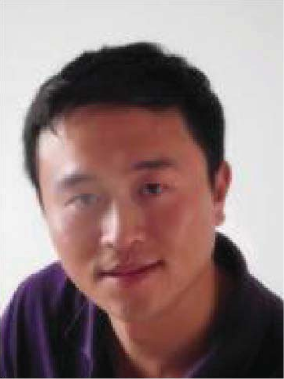}}]{Jun Yu}(M'13) received the B.Eng. and Ph.D. degrees from Zhejiang University, Zhejiang, China.
He is currently a Professor with the School of Computer Science and Technology, Hangzhou Dianzi University, Hangzhou, China. He was an Associate Professor with the School of Information Science and Technology, Xiamen University, Xiamen, China. From 2009 to 2011, he was with Nanyang Technological University, Singapore. From 2012 to 2013, he was a Visiting Researcher at Microsoft Research Asia (MSRA). Over the past years, his research interests have included multimedia analysis, machine learning, and image processing. He has authored or coauthored more than 60 scientific articles. In 2017 Prof. Yu received the IEEE SPS Best Paper Award.

Prof. Yu has (co-)chaired several special sessions, invited sessions, and workshops. He served as a program committee member or reviewer of top conferences and prestigious journals. He is a Professional Member of the Association for Computing Machinery (ACM) and the China Computer Federation (CCF).
\end{IEEEbiography}
\vspace{-10pt}
\begin{IEEEbiography}[{\includegraphics[width=1.2in,height=1.25in,clip,keepaspectratio]
{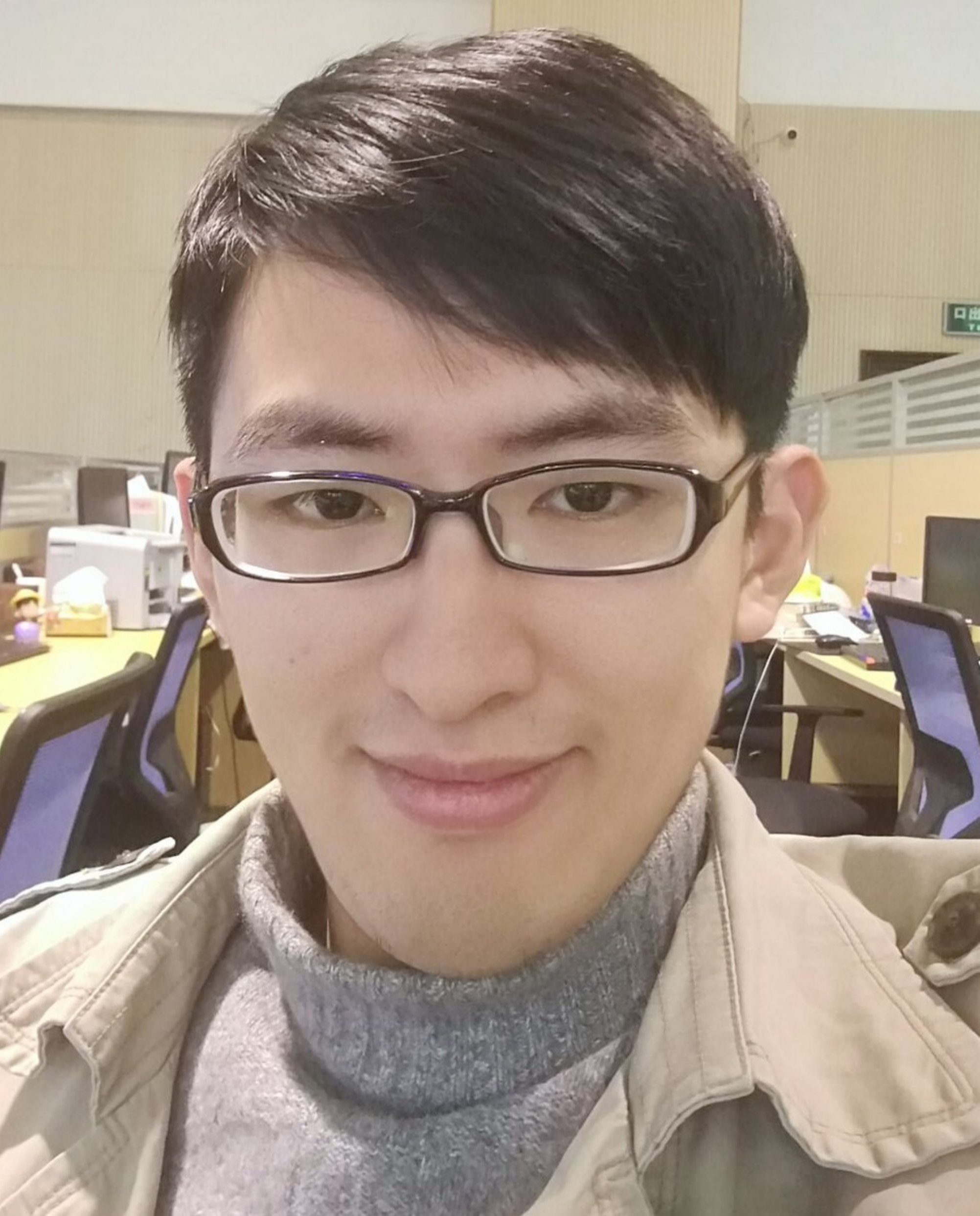}}]{Chenchao Xiang}
received the B.Eng. degree from the School of Management, Hangzhou Dianzi University, Hangzhou, China, in 2016. He is currently pursuing the M.Eng. degree with the College of Computer Science and Technology. His current research interests include multimodal analysis, computer vision and machine learning.
\end{IEEEbiography}
\vspace{-10pt}
\begin{IEEEbiography}[{\includegraphics[width=1.2in,height=1.25in,clip,keepaspectratio] {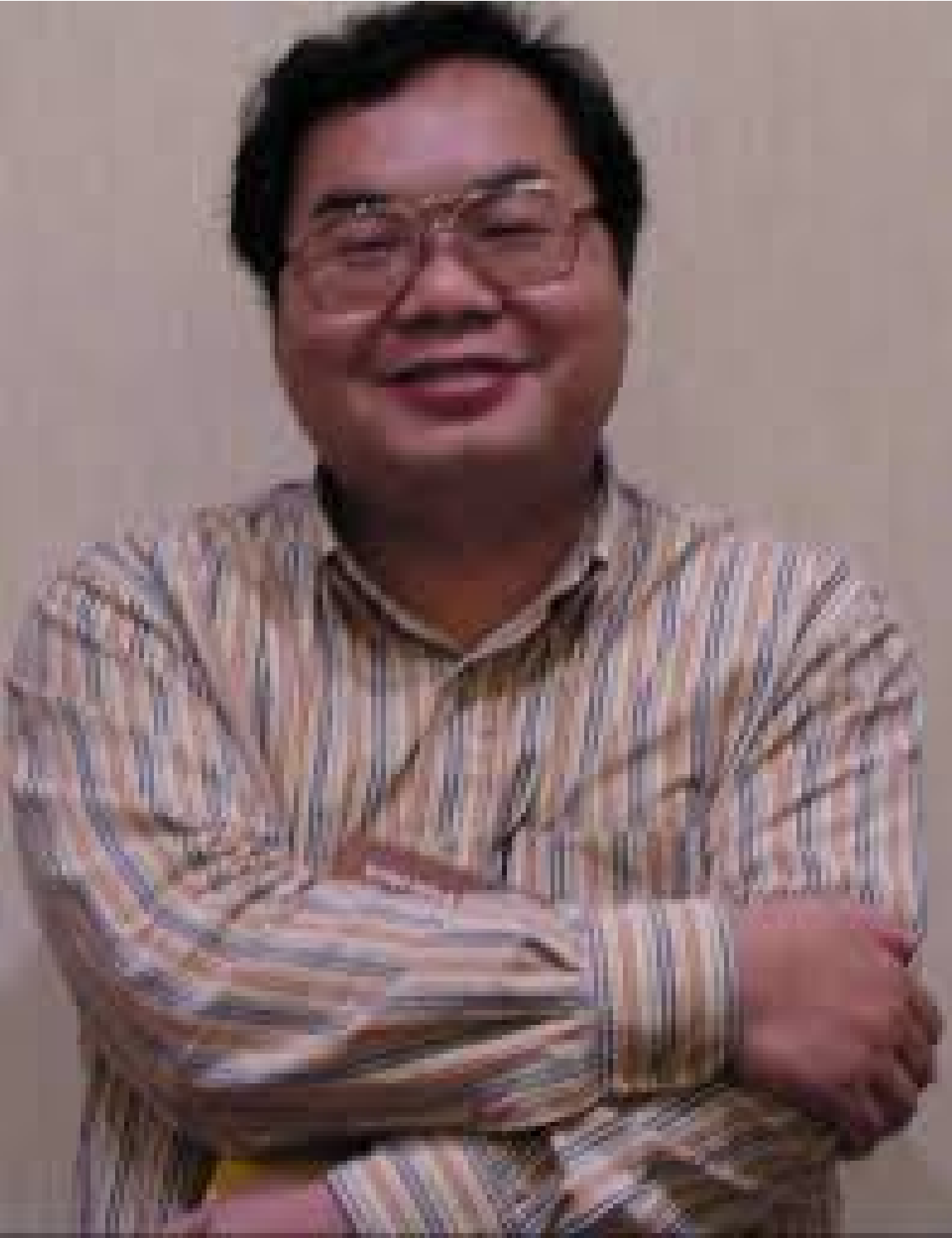}}] {Jianping Fan} is a professor at UNC-Charlotte. He received his MS degree in theory physics from Northwestern University, Xian, China in 1994 and his PhD degree in optical storage and computer science from Shanghai Institute of Optics and Fine Mechanics, Chinese Academy of Sciences, Shanghai, China, in 1997. He was a Postdoc Researcher at Fudan University, Shanghai, China, during 1997-1998. From 1998 to 1999, he was a Researcher with Japan Society of Promotion of Science (JSPS), Department of Information System Engineering, Osaka University, Osaka, Japan. From 1999 to 2001, he was a Postdoc Researcher in the Department of Computer
Science, Purdue University, West Lafayette, IN. His research interests include image/video privacy protection, automatic image/video understanding, and large-scale deep learning.
\end{IEEEbiography}
\vspace{-10pt}
\begin{IEEEbiography}[{\includegraphics[width=1.0in,height=1.2in,clip,keepaspectratio]
{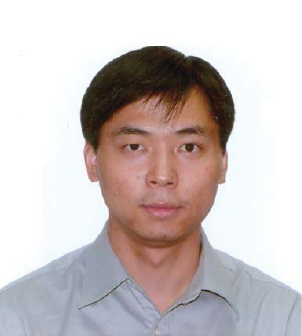}}]{Dacheng Tao}(F'15) is Professor of Computer Science and ARC Laureate Fellow in the School of Information Technologies and the Faculty of Engineering and Information Technologies, and the Inaugural Director of the UBTECH Sydney Artificial Intelligence Centre, at the University of Sydney. He mainly applies statistics and mathematics to Artificial Intelligence and Data Science. His research interests spread across computer vision, data science, image processing, machine learning, and video surveillance. His research results have expounded in one monograph and 500+ publications at prestigious journals and prominent conferences, such as IEEE T-PAMI, T-NNLS, T-IP, JMLR, IJCV, NIPS, ICML, CVPR, ICCV, ECCV, ICDM; and ACM SIGKDD, with several best paper awards, such as the best theory/algorithm paper runner up award in IEEE ICDM07, the best student paper award in IEEE ICDM13, the distinguished student paper award in the 2017 IJCAI, the 2014 ICDM 10-year highest-impact paper award, and the 2017 IEEE Signal Processing Society Best Paper Award. He received the 2015 Australian Scopus-Eureka Prize, the 2015 ACS Gold Disruptor Award and the 2015 UTS Vice-Chancellor's Medal for Exceptional Research. He is a Fellow of the IEEE, AAAS, OSA, IAPR and SPIE.
\end{IEEEbiography}





\end{document}